\documentclass{article}

\usepackage{arxiv}
\usepackage{amsmath}
\usepackage[utf8]{inputenc} 
\usepackage[T1]{fontenc}    
\usepackage{hyperref}       
\usepackage{url}            
\usepackage{booktabs}       
\usepackage{amsfonts}       
\usepackage{nicefrac}       
\usepackage{microtype}      
\usepackage{lipsum}		
\usepackage{graphicx}
\usepackage[square,numbers]{natbib}
\usepackage{doi}
\usepackage{xcolor}
\usepackage[section]{placeins}
\usepackage{multirow}
\usepackage{rotating} 
\usepackage{color}
\newcommand\revs[1]{\textcolor{black}{#1}}

\title{Towards Robust Surrogate Models: Benchmarking Machine Learning Approaches to Expediting Phase Field Simulations of Brittle Fracture}

	\author{ \href{https://orcid.org/0009-0008-3262-7533}{\includegraphics[scale=0.06]{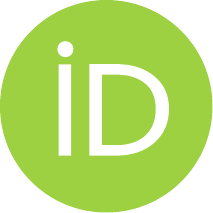}\hspace{1mm}Erfan~Hamdi} \\
	Department of Mechanical Engineering\\
	Boston University\\
	Boston, MA 02215 \\
	\texttt{erfan@bu.edu} \\
 	\And
   \href{https://orcid.org/0000-0001-8099-3468}{\includegraphics[scale=0.06]{orcid.pdf}\hspace{1mm}Emma~Lejeune$^1$}\\
	Department of Mechanical Engineering\\
	Boston University\\
	Boston, MA 02215 \\
	\texttt{elejeune@bu.edu} \\
}

\renewcommand{\shorttitle}{Benchmark Dataset for Solid Mechanics: Phase Field Modeling of Brittle Fracture}

\hypersetup{
pdftitle={Towards Robust Surrogate Models: Benchmarking Machine Learning Approaches to Expediting Phase Field Simulations of Brittle Fracture},
pdfauthor={Erfan~Hamdi,Emma~Lejeune},
}

\begin{document}
\maketitle

\footnotetext[1]{corresponding author}

\begin{abstract}
Data-driven approaches have the potential to make modeling complex, nonlinear physical phenomena significantly more computationally tractable. For example, computational modeling of fracture is a core challenge where machine learning techniques have the potential to provide a much needed speedup that would enable progress in areas such as multi-scale modeling and uncertainty quantification. 
Currently, phase field modeling (PFM) of fracture is one such approach that offers a convenient variational formulation to model crack nucleation, branching and propagation. To date, machine learning techniques have shown promise in approximating PFM simulations. 
\revs{While standard fracture benchmarks represent realistic scenarios frequently observed in practice, they typically do not provide sufficiently challenging tests for data-driven methods.}
To address this gap, we introduce a challenging dataset based on PFM simulations designed to benchmark and advance ML methods for fracture modeling. This dataset includes three energy decomposition methods, two boundary conditions, and 1,000 random initial crack configurations for a total of 6,000 simulations. Each sample contains 100 time steps capturing the temporal evolution of the crack field. 
Alongside this dataset, we also implement and evaluate Physics Informed Neural Networks (PINN), Fourier Neural Operators (FNO) and UNet models as baselines, and explore the impact of ensembling strategies on prediction accuracy. With this combination of our dataset and baseline models drawn from the literature we aim to provide a standardized and challenging benchmark for evaluating machine learning approaches to solid mechanics.  Our results highlight both the promise and limitations of popular current models, and demonstrate the utility of this dataset as a testbed for advancing machine learning in fracture mechanics research.
\end{abstract}

\section{Introduction}
\label{sec:intro}
In recent years, data-driven modeling has gained significant traction across science and engineering due to its ability to approximate complex physical systems at a fraction of the computational cost of traditional high fidelity simulations~\cite{forrester2009recent, lu2023drips} for autonomous and non-autonomous systems where the input space varies with time~\cite{lu2024data}.
\revs{This advantage is critical for challenges like uncertainty quantification, particularly in energy storage devices such as redox flow batteries~\cite{liu2020data}. In these systems, complex phenomena like fracture and crack propagation contribute to capacity fade and reduced lifespan~\cite{boyce2022cracking}. The design process for these complex systems involves studying coupled phenomena across multiple scales, making the use of traditional computational methods prohibitively expensive. Data-driven approaches, therefore, offer a promising solution for achieving the speedups required for effective simulation~\cite{mo2019deep}.}
By extracting patterns from data and learning compact representations, neural network based approximations can offer fast and flexible alternatives to conventional solvers. A wide range of architectures have been proposed to introduce inductive biases relevant to the underlying structure of physical problems. For instance, convolutional neural networks (CNNs) have been applied to structured grids~\cite{bhatnagar2019prediction} while graph neural networks (GNNs) have been applied to problems defined on unstructured meshes and irregular underlying structures~\cite{prachaseree2022learning, zhao2024review}. To better integrate governing physics into the learning process, physics informed neural networks (PINNs)~\cite{raissi2019physics} were introduced. These models use neural networks to parameterize the solution of partial differential equations (PDEs) and incorporate the residuals of the governing equations directly into the loss function. PINNs have been applied to a broad range of problems~\cite{mao2020physics, sahli2020physics, haghighat2021physics, fang2019deep, misyris2020physics}, but face scalability challenges as they typically need to be retrained for every change in initial or boundary condition. To overcome these limitations, operator learning approaches have emerged. These methods aim to learn mappings between infinite dimensional input and output function spaces, enabling generalization across different input space parameters such as initial and boundary conditions. Both purely data-driven and physics constrained variants have been proposed within this framework achieving state of the art performance across a wide range of scientific and engineering domains~\cite{lu2019deeponet, li2020neural, patel2021physics, wang2021learning, kiyani2025predicting, goswami2023physics}.

However, despite these advances, neural network based solutions continue to face major challenges. For example, the models mentioned previously, apply the governing physical laws as soft penalties within the loss function rather than enforcing them strictly. As a result, the learned solutions may violate the underlying physics especially in regions with sparse data or high complexity~\cite{utkarsh2025physics, hansen2023learning, wang2021understanding, berrone2023enforcing}. 
Additionally, many models suffer from a phenomenon known as spectral bias which is the tendency of the model to learn low frequency (smooth) components of the solution first while struggling to capture high frequency features~\cite{rahaman2019spectral, khodakarami2025mitigating}. This makes it challenging for these models to accurately capture regions with sharp gradients, such as discontinuities or localized damage. This behavior is particularly problematic for modeling fracture where small scale features such as cracks play a critical role in determining the system's overall response. 
Other persistent issues include difficulty of optimizing the neural networks in a non-convex loss landscape~\cite{dauphin2014identifying, pascanu2014saddle, kiyani2025optimizer} and sensitivity to initialization or training dynamics~\cite{frankle2018lottery, glorot2010understanding, bouthillier2021accounting}. 
Together, these limitations make it difficult to develop robust, generalizable and physically faithful models highlighting the need for structured, high fidelity benchmark datasets to systematically study and stress test model capabilities such as the Mechanical MNIST collection~\cite{lejeune2020mechanical, nguyen2023mechanical, kobeissi2022mechanical} which presented accessible benchmark datasets for mechanical engineering inspired from the computer vision research community or other benchmarks that target broader scientific meta modeling benchmarks such as the Well~\cite{ohana2024well} or PDEBench~\cite{takamoto2022pdebench}. 

Inspired by these challenges, we have chosen fracture as the starting point for our dataset.
\revs{Fracture and crack propagation is a particularly critical failure mechanism in real engineering systems. For example, in energy storage devices such as lithium-ion batteries,
imperfections such as pre-existing microcracks can propagate through an electrode and limit functionality.~\cite{boyce2022cracking, woodford2010electrochemical, klinsmann2016modeling}}.
There are many different ways to model brittle fracture~\cite{sukumar2000extended, talebi2013molecular, lorentz1999variational}. Among these methods, Phase field modeling (PFM) of brittle fracture is a popular modern option that has been explored recently within the ML literature~\cite{goswami2020transfer, manav2024phase}. Notably fracture problems are inherently characterized by discontinuities, sharp gradients, and topological changes such as crack nucleation, branching, and coalescence, which, as mentioned, are particularly difficult for data-driven models to capture accurately. At the same time, the coupled governing equations of the PFM provide a well defined and physically grounded framework with intrinsic non-convexity~\cite{gerasimov2020stochastic} making it a natural setting to embed physics into neural networks and rigorously evaluate their robustness and convergence behavior under complex physical settings. PFMs have also been explored in the context of stochastic modeling~\cite{gerasimov2020stochastic} and uncertainty quantification~\cite{zhang2024representing}, and are notorious for being computationally expensive~\cite{wu2020phase, miehe2010phase, ambati2015review}.
Different attempts have been made by researchers to offer a data-driven approximation to the solution of phase field models of brittle fracture~\cite{goswami2020adaptive, goswami2022physics, manav2024phase, ghaffari2023deep, kiyani2025predicting}. However many of these studies have ultimately evaluated ML models on relatively simple benchmark problems. Without a rigorous exploration of more complex and diverse fracture scenarios, it remains unclear whether these ML models are truly able to capture the subtleties of fracture prediction.

\begin{figure}[!htb]
    \centering
    \includegraphics[width=\linewidth]{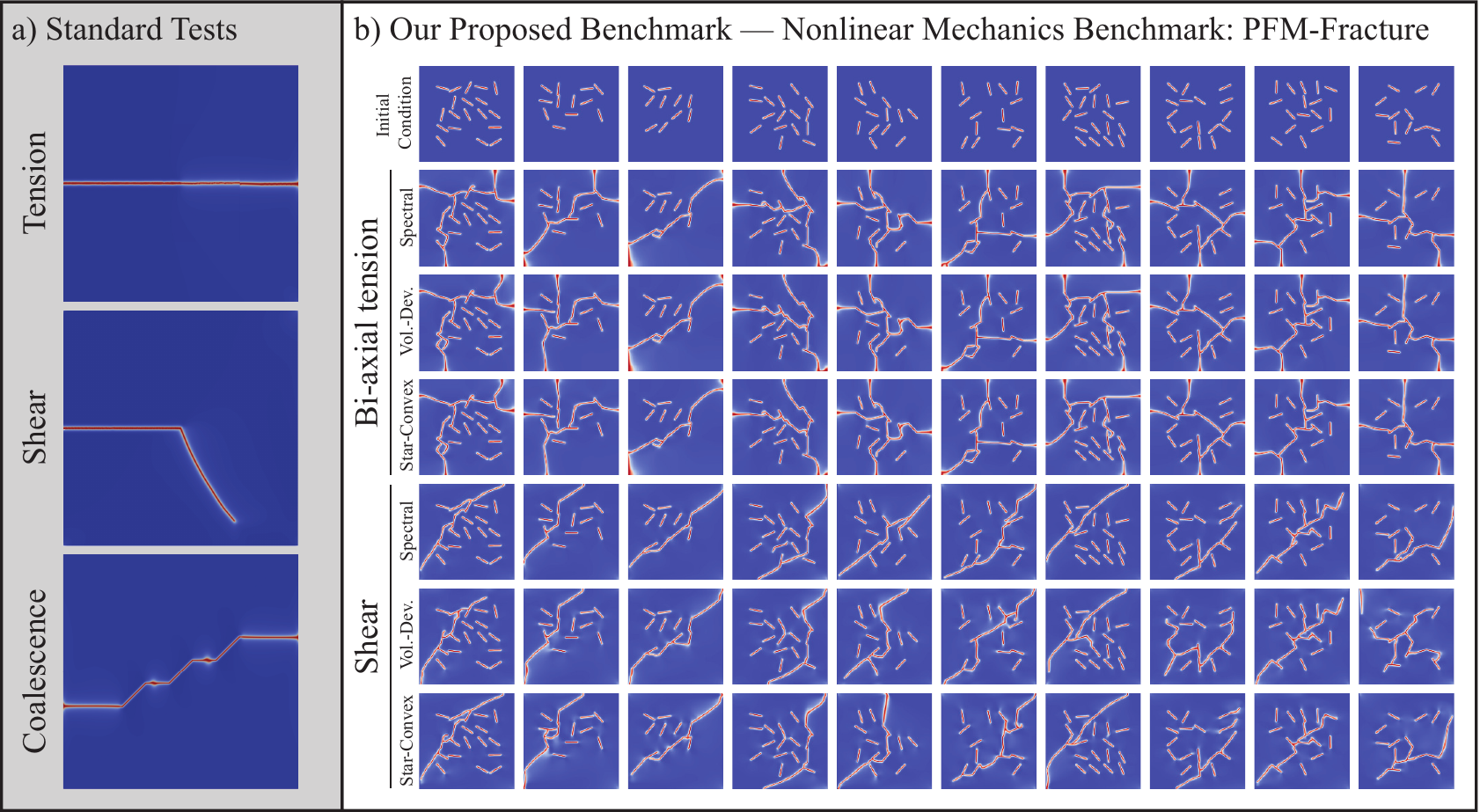}
    \caption{Comparison between approaches to evaluating surrogate model efficacy: a) Standard examples from the literature such as tension, shear, and crack coalescence~\cite{goswami2020adaptive, manav2024phase}, b) Our proposed benchmark dataset designed to provide a diverse set of challenges across initial conditions, load conditions, and energy decomposition methods.}
    \label{fig:intro}
\end{figure}

In this work our primary goal is to develop a challenging dataset that enables a deeper exploration of the subtleties of surrogate modeling with a particular focus on phase field modeling of brittle fracture. As can be seen in Fig.~\ref{fig:intro} this dataset is designed not only to reflect the complex nature of fracture and crack propagation under various boundary conditions and energy decompositions, but also to serve as a robust benchmark for evaluating the performance of data-driven surrogate models. Specifically, we implement and assess the performance of several baseline neural network based models including Physics informed neural networks (PINNs), UNets, and Fourier Neural Operators (FNOs) in order to better understand their strengths and limitations in modeling fracture. 
A further objective of this work is to evaluate the impact of three ensembling strategies (hard voting, soft voting and stacking) on the performance of these models, and to assess their effectiveness in improving predictive accuracy and robustness.
\revs{This is motivated by the fact that a significant challenge in training surrogate models on a finite number of data points is the introduction of epistemic uncertainty due to approximation errors~\cite{tripathy2018deep}. While many approaches exist for quantifying uncertainty in neural networks~\cite{tripathy2018deep, graves2011practical}, ensembling methods based on random initialization offer a simple yet effective strategy. By exploring different regions of the non-convex loss landscape, ensembles can improve the predictive performance of the neural networks and provide a practical method for uncertainty quantification~\cite{yang2022scalable, fort2019deep, psaros2023uncertainty}.}
This dataset extends our previous contributions to benchmark dataset development in mechanics, including the Mechanical MNIST dataset collection~\cite{lejeune2020mechanical}. Specifically, it builds upon the Mechanical MNIST Crack Path dataset~\cite{mohammadzadeh2021mechanical} by introducing a more advanced and more broadly relevant dataset. We refer to this new dataset, "PFM-Fracture" subset of our new "Nonlinear Mechanics Benchmark" dataset which consists of high-resolution finite element simulations of brittle fracture using PFM applied to domains with randomly initialized cracks.

The remainder of this paper is organized as follows. First in Section ~\ref{sec:pfm}, we will start by discussing phase field modeling of brittle fracture. Next, in Section~\ref{sec:dataset_spec}, we will present the specifications of the dataset that was created. After that, we will cover the details of the implemented baseline models and ensembling methods in Sections~\ref{sec:baseline} and~\ref{sec:ensembling-methods}. Then, in Section~\ref{sec:results}, we will present the results of training them on the dataset as well as the outcome of applying ensembling methods to improve model predictions. We will compare the performance of these baseline models with each other in Section~\ref{sec:comparisons}, and we will briefly go over the limitations of the metrics used in our training process and evaluation of machine learning models. Finally in Section~\ref{sec:conclusion}, we present the conclusions and outlook of this work.

\section{Methods}
\label{sec:methods}
We begin in Section~\ref{sec:pfm} by introducing the phase field fracture formulation for brittle materials and the three different energy decomposition methods implemented within our dataset.
Then, in Section~\ref{sec:dataset_spec}, we describe in detail the initial and boundary conditions used in our dataset, and the dataset file structure. Finally, in Section~\ref{sec:baseline}, we define the baseline machine learning models associated with this manuscript. We also note briefly that all code associated with the generation of this dataset and the implementation of these baseline models is available on GitHub, with a link provided in Section~\ref{sec:additional_info}.

\subsection{Phase Field Model Preliminaries}
\label{sec:pfm}
\subsubsection{Phase Field Modeling of Fracture}
\label{subsec:pfm}
Fracture is a common failure mechanism for engineering materials and structures~\cite{anderson2005fracture}. From both a theoretical and computational standpoint, fracture remains one of the most challenging phenomena to model~\cite{wu2020phase}. Early continuum mechanics approaches began with Griffith's theory~\cite{griffith1921vi}, which describes fracture as a competition between crack surface energy and elastic energy, and continued with the development of theory such as linear elastic fracture mechanics (LEFM) by Irwin~\cite{irwin1957analysis}, and the cohesive zone model (CZM) by Dugdale~\cite{dugdale1960yielding}. However, these methods rely on external criteria to predict crack nucleation and propagation, which prompted the search for alternative approaches, particularly approaches that would be amenable to simulation~\cite{wu2020phase}. One promising and recently popular method with broad adoption in the literature is phase field modeling (PFM) of fracture~\cite{kuhn2010continuum}.
Phase field modeling of brittle fracture offers a variational approach to modeling fracture and crack propagation by transforming the problem into the minimization of a potential energy function. This is achieved through regularization of the crack via a scalar field, called the phase field, using a characteristic length $l_0$ that defines the extent of regularization~\cite{miehe2010phase}. In the introduced phase field, the cracked regions smoothly transition from a scalar value of $\phi = 1$ to the intact regions with a scalar value of $\phi = 0$.

\begin{figure}[!htb]
    \centering
    \includegraphics[width=\linewidth]{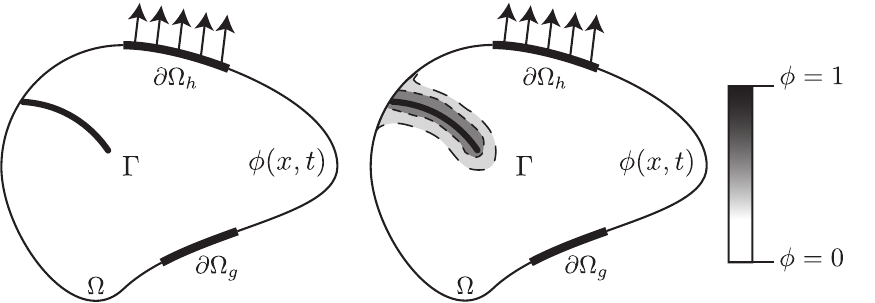}
    \caption{A general schematic of a mechanical system with a) the real sharp crack surface $\Gamma$ inside a domain $\Omega$ with boundary conditions, and b) the crack approximation via the scalar phase field with $\phi = 1$ in the cracked region and $\phi=0$ in the intact region with a diffuse transition between the regions.}
    \label{fig:pfm-intro}
\end{figure}

Miehe et al.~\cite{miehe2010phase} regularized the crack using a crack surface density function $\gamma(\phi, \nabla \phi)$. In this way, the original variational formulation of Griffith~\cite{griffith1921vi}, which required the integration of crack surface energy over the unknown crack surface $\Gamma$, was transformed into an integration over the entire domain $\Omega \in R^d$ with Dirichlet boundary conditions on $\partial\Omega_g \subseteq \partial\Omega$ and Neumann boundary conditions on $\partial\Omega_h \subseteq \partial\Omega$ as shown in Fig.~\ref{fig:pfm-intro}. With this definition, we can define total energy potential $\mathcal{E}$ as

\begin{equation}
    \mathcal{E}(\mathbf{u}, \phi) = \Psi_s + \Psi_c - \mathcal{P} 
                        = \int_{\Omega} \psi_0(\mathbf{\epsilon}(\mathbf{u}))\mathrm{d}V 
    + \int_{\Gamma} G_c \mathrm{d}A - \int_\Omega \mathbf{b}\cdot\mathbf{u}\mathrm{d}V + \int_{\partial\Omega}\mathbf{t}\cdot\mathbf{u}\mathrm{d}A
\end{equation}

where $\mathcal{E}$ is a function of the displacement $\mathbf{u}$, the phase field $\phi$, and the potential energy functional $\mathcal{P}$ of external forces is a function of distributed body force $\mathbf{b}$ and traction on the boundaries $\mathbf{t}$.
The first term, $\Psi_s$, is defined as the strain energy potential and $\psi_0$ is the strain energy density function as a function of infinitesimal strain tensor $\mathbf{\epsilon}(x)$. We define  $\mathbf{\epsilon}(x)$ as:

\begin{equation}
    \mathbf{\epsilon}(\mathbf{x}) := \nabla^s \mathbf{u} 
\end{equation}

where $\nabla^s (\cdot)$ is the symmetric gradient operator. Assuming a linear elastic material, the strain energy density function is then defined as:

\begin{equation}
    \psi_0 = \frac{1}{2} \lambda \operatorname{tr}^2(\mathbf{\epsilon})+\mu\operatorname{tr}(\mathbf{\epsilon}^2)
\end{equation}

with $\lambda$ and $\mu$ as the Lam\'e constants.
The dissipated surface energy is defined as $\Psi_c$.
The surface energy term can be regularized as:

\begin{equation}
\Psi_c = \int_{\Gamma} G_c\mathrm{d}A \approx \int_\Omega G_c \gamma(\phi, \nabla \phi)\mathrm{d}A
\end{equation}

where $G_c$ is the critical energy release rate of the material. The crack surface density function is defined as:

\begin{equation}
\label{eq:gamma}
    \gamma(\phi, \nabla \phi) = \frac{1}{2} 
\left[ \frac{1}{l_0} \phi^2 + l_0 (\nabla \phi \cdot \nabla \phi) \right]
\end{equation}

where $l_0$ is the characteristic length. Literature suggests that $l_0$ should be treated as a material property that governs the regularization of a sharp crack~\cite{amor2009regularized, wu2020phase}. As $l_0 \rightarrow 0$ the solution converges to the original fracture problem; however, capturing a small characteristic length requires a highly refined mesh in the transition zone leading to a significant computational cost~\cite{gerasimov2020stochastic}.
After introducing the regularization to the crack surface energy term, the bulk strain energy term is now defined as:

\begin{equation}
\label{eq:psi_strain}
    \Psi_s (\mathbf{u}, \phi) = \int_{\Omega} g(\phi)\psi_0(\mathbf{\epsilon}(\mathbf{u}))\mathrm{d}V
\end{equation}

where $g(\phi)$ refers to the energetic degradation function.
With this background, the total energy functional can be written as:

\begin{equation}
\label{eq:total_func}
    \mathcal{E}(\mathbf{u}, \phi) = 
\int_{\Omega} g(\phi) \psi_0(\mathbf{\epsilon}(\mathbf{u})) \, \mathrm{d}V 
+ \int_{\Omega} G_c \frac{1}{2} 
\left[ \frac{1}{l_0} \phi^2 + l_0 (\nabla \phi \cdot \nabla \phi) \right] 
\, \mathrm{d}V 
- \int_\Omega \mathbf{b}\cdot\mathbf{u}\mathrm{d}V + \int_{\partial\Omega}\mathbf{t}\cdot\mathbf{u}\mathrm{d}A \, .
\end{equation}

Various variational and non-variational approaches have been proposed in the literature. In variational formulations, the stress field and phase field evolution equations can be derived from Eq.\ref{eq:total_func}, as seen in~\cite{bourdin2008variational, miehe2010phase}. However, these formulations lead to nonlinear equilibrium equations, which are computationally expensive to solve. To address this issue, hybrid formulations have been introduced~\cite{ambati2015review, wu2017unified}, incorporating a linear displacement subproblem to reduce the computational cost. Given the complexity of each simulation sample—particularly due to the need to resolve crack coalescence and the irregularity of initial conditions—we have implemented the hybrid phase field formulation proposed by Ambati et al.~\cite{ambati2015review}, which maintains a linear momentum equation to reduce computational costs.
With this approach, the linear momentum equation is written as follows:

\begin{equation}
\left\{ \begin{aligned} 
  \sigma(\mathbf{u}, \phi) :=& (1-\phi)^2\frac{\partial\psi_0(\mathbf{\epsilon})}{\partial \mathbf{\epsilon}},\\
  -l_0^2\Delta \phi + \phi =& \frac{2l_0}{G_c}(1-\phi)\mathcal{H}^+,\\
  \mathcal{H}^+(x, t) :=& \max_{\tau \in [0, t]} \psi_0^+ \big( \mathbf{\epsilon}(x, \tau) \big)
\end{aligned} \right.
\end{equation}

where the history field variable $\mathcal{H}^+$ is adopted from~\cite{miehe2010phase} to enforce the irreversibility of the crack phase field evolution where $\psi_0^+$ is the tensile component of elastic energy density.
Each of the displacement field and phase field functional are minimized iteratively in an alternating minimization framework, which offers a robust but slow convergence~\cite{wu2020phase}. An important feature of Eq.\ref{eq:total_func} is that it is convex with respect to  $\mathbf{u}$ and $\phi$, but is not convex with respect to both~\cite{gerasimov2020stochastic}. This further necessitates the use of an alternating minimization method for solving the PDEs~\cite{wu2020phase}.
To enforce tensile-compressive asymmetry in the fracture response, the elastic energy density function must be decomposed into tensile and compressive parts where only the positive component should be degraded using the energy degradation function $g(\phi) = (1 - \phi)^2$ as in Eq.\ref{eq:decompose}. With this approach, $\psi_0(\mathbf{\epsilon}(\mathbf{u}))$ is written as:

\begin{equation}
    \psi_0(\mathbf{\epsilon}(\mathbf{u})) = g(\phi)\psi_0^+(\mathbf{\epsilon}(\mathbf{u})) +  \psi_0^-(\mathbf{\epsilon}(\mathbf{u})).
\label{eq:decompose}
\end{equation}

In this work, we implement three different energy decompositions methods: volumetric-deviatoric decomposition proposed by Amor et al.~\cite{amor2009regularized}, spectral decomposition proposed by Miehe et al.~\cite{miehe2010phase}, and the star-convex decomposition proposed by Vicentini et al.~\cite{vicentini2024energy}. Each of these decompositions will be discussed further in Section~\ref{sec:energy-decomp}.
We also briefly note that, though we do not cover them here, there have been several different approaches in the literature including non-variational phase field modeling that will allow for crack healing and nucleation under compression, and better prediction of crack nucleation in the absence of large pre-existing cracks by utilizing the material strength surface instead of energy splitting methods~\cite{kumar2018fracture, kamarei2024poker, liu2025emergence, senthilnathan2025construction}.

\subsubsection{Energy Decomposition Methods}
\label{sec:energy-decomp} 
In order to best represent the state of the art in phase field fracture simulations, our dataset contains simulation results from three different energy decomposition methods. The first two, spectral decomposition~\cite{miehe2010phase} and volumetric-deviatoric decomposition~\cite{amor2009regularized}, have a long history of adoption by the community. However, multiple different alternative approaches to energy decomposition have been suggested in the literature~\cite{freddi2010regularized, de2022nucleation, vajari2023investigation, tang2019phase}. Recently, Vicentini et al.~\cite{vicentini2024energy} offered a modification to the volumetric-deviatoric decomposition utilizing a parameter $\gamma^*$ to address the shortcomings of other decomposition methods in modeling crack nucleation and propagation. We provide an implementation of this method, the star-convex method, as an additional example of a simple tunable method allowing the calibration of tensile, compressive, and shear strength of a material under multi-axial loading.

\subsubsection*{Spectral Decomposition}
An energy decomposition method, suggested by Miehe et al.~\cite{miehe2010phase}, decomposes the strain into tensile and compressive parts by using the eigenvalues and eigenvectors of the strain tensor written as:

\begin{align}
    \mathbf{\epsilon}^+ &= \sum_i \langle \mathbf{\epsilon}_i \rangle_+ e_i \otimes e_i \\
    \mathbf{\epsilon}^- &= \sum_i \langle \mathbf{\epsilon}_i \rangle_- e_i \otimes e_i
\end{align}

where $\mathbf{\epsilon}_i$ is the eigenvalue, $e_i$ is the corresponding eigenvector, and $\otimes$ refers to the dyadic product.
This decomposition leads to the expression:

\begin{align}
    \psi_0^+(\mathbf{\epsilon}) &= \frac{1}{2} \lambda_0 \langle \operatorname{tr}(\mathbf{\epsilon}) \rangle^2 
+ \mu_0 \mathbf{\epsilon}^+ : \mathbf{\epsilon}^+ \,\\
    \psi_0^-(\mathbf{\epsilon}) &= \frac{1}{2} \lambda_0 \langle -\operatorname{tr}(\mathbf{\epsilon}) \rangle^2 
+ \mu_0 \mathbf{\epsilon}^- : \mathbf{\epsilon}^- 
\end{align}

Though this decomposition is computationally more intensive, it better models compression-dominated loadings, where all three principal strains are negative when compared to the volumetric-deviatoric decomposition approach~\cite{van2020strain, wu2020phase}.

\subsubsection*{Volumetric-Deviatoric Decomposition}
This decomposition method introduced by Amor et al.~\cite{amor2009regularized} decomposes the strain tensor into a deviatoric $\mathbf{\epsilon}_D$ term and a volumetric $\mathbf{\epsilon}_V$ term as follows: 

\begin{align}
\mathbf{\epsilon} &= \mathbf{\epsilon}_V + \mathbf{\epsilon}_D \\ 
\mathbf{\epsilon}_V &= \frac{1}{3} \operatorname{tr}(\mathbf{\epsilon}) \mathbf{1}\\
\mathbf{\epsilon}_D &= \mathbf{\epsilon} - \frac{1}{3} \operatorname{tr}(\mathbf{\epsilon}) \mathbf{1}
\end{align}

The resulting strain energy density potential is defined as:

\begin{align}
    \psi_0^+(\mathbf{\epsilon}) &= \frac{1}{2} \kappa_0 \langle \operatorname{tr}(\mathbf{\epsilon}) \rangle^2 + \mu_0 \mathbf{\epsilon}_D : \mathbf{\epsilon}_D
\\
\psi_0^-(\mathbf{\epsilon}) &= \frac{1}{2} \kappa_0 \langle -\operatorname{tr}(\mathbf{\epsilon}) \rangle^2 
\end{align}

where the $\langle \rangle$ is defined as $\langle a \rangle := \max\{a, 0\}$, $\kappa_0 = \lambda + 2\mu/3$ denotes the bulk modulus, and $\mathbf{1}$ is the identity tensor. 
This approach offers a straightforward decomposition method that effectively captures key requirements for defining a new energy decomposition in the phase field method, as outlined in~\cite{vicentini2024energy}, while ensuring a feasible computational cost.

\subsubsection*{Star-Convex Decomposition}
We now introduce the recently proposed star-convex energy decomposition method, which can be viewed as a modification of the volumetric-deviatoric approach aimed at enhancing its flexibility by allowing for calibration of compressive $\sigma_{e}^-$ and tensile strength $\sigma_{e}^+$. In this approach, a $\gamma^*$ term is introduced to allow for a residual stress related to volume contraction $\langle -\operatorname{tr}(\mathbf{\epsilon}) \rangle$, and the strain energy is then written as:

\begin{align}
    &\psi_0^+(\mathbf{\epsilon}) =  \frac{1}{2} \kappa_0 
\left( \langle \operatorname{tr}(\mathbf{\epsilon}) \rangle^2 
- \gamma^\star \langle -\operatorname{tr}(\mathbf{\epsilon}) \rangle^2 \right) + \mu_0 \mathbf{\epsilon}_{D} : \mathbf{\epsilon}_D
\\
&\psi_0^-(\mathbf{\epsilon}) = (1 + \gamma^\star) 
\frac{1}{2} \kappa_0 
\langle -\operatorname{tr}(\mathbf{\epsilon}) \rangle^2\\
\end{align}

where $\gamma^* = \sigma_{e}^-/\sigma_{e}^+$. We note briefly that in this decomposition method, the original volumetric-deviatoric decomposition can be reconstructed by setting the term $\gamma^* = 0$. In this work, we select $\gamma^* = 5$ to match the original reference.
\begin{figure}[!h]
    \centering
    \includegraphics[width=\linewidth]{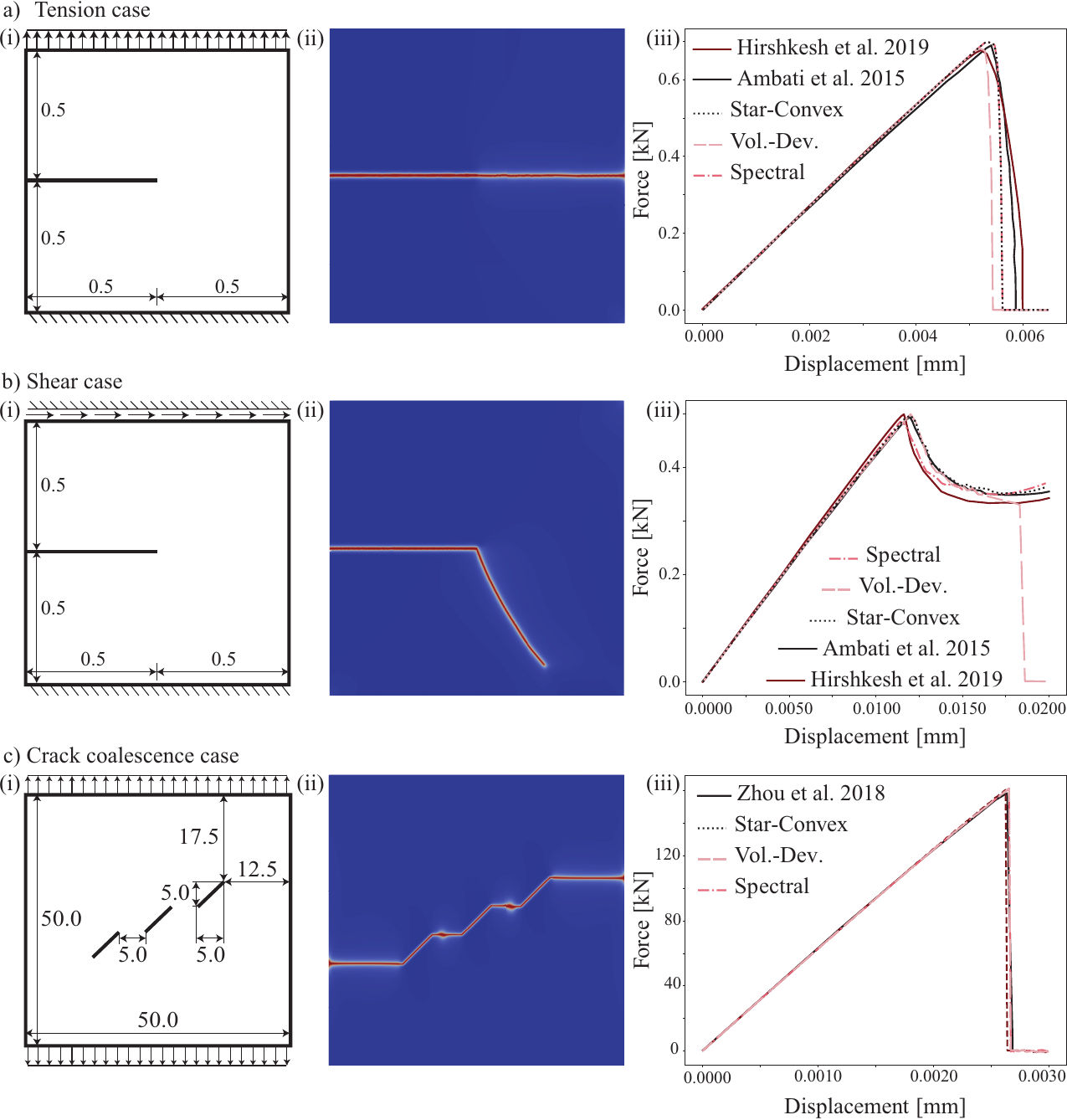}
    \caption{Geometrical description and boundary conditions of the three benchmark cases and the resulting crack path and force displacement curves to validate the implemented phase field model. Here we show: a) the Tension case, b) the Shear case and c) the crack coalescence case under uniaxial tension. The crack patterns for the spectral decomposition is depicted in (ii). The force displacement curve for all three of our energy decomposition approaches are reported in (iii) alongside results from the literature.}
    \label{fig:benchmarks}
\end{figure}
\subsubsection{Benchmark Validation Problems}
\label{sec:validation}
When introducing a new phase field formulation and validating a phase field model implementation, it is standard to report performance on multiple benchmark examples~\cite{ambati2015review, wu2018length, kumar2022revisited}.
Many different benchmark cases have been suggested in the phase field fracture literature, some with the aim of examining the effect of different modeling assumptions, and others with the aim of ensuring correct model implementation. At present, there are published benchmarks relevant to quasi-static fracture~\cite{ambati2015review},  dynamic fracture~\cite{borden2012phase}, hydraulic fracture~\cite{mikelic2015phase, zhou2018phasePoro}, crack coalescence~\cite{zhou2018phase}, multi-axial loading, tensile loading, compressive loading, and shear loading~\cite{zhuang2022phase}. 

However, despite the fact that comparing to standard benchmarks is common practice, in our analysis of the literature we found that it was often quite challenging to reproduce the benchmark results reported by others. Critically, phase field fracture approaches are sensitive to multiple modeling parameters, including basic mesh structure, characteristic length, convergence criteria, and the strategy for modeling pre-existing cracks~\cite{gerasimov2020stochastic, wu2020phase}. 
And, because sharing software and implementation code are not standard practice, it is especially difficult to isolate sources of discrepancy. To this end, we make a note of our implementation choices in Section~\ref{sec:dataset_spec}, and share all of the code underlying our implementation of the models shown in this Section.
To validate our implementation of the three approaches outlined in Section~\ref{sec:energy-decomp} under different types of loading and boundary conditions, we compare their performance to each other and to the literature on three standard benchmark problems. These popular benchmark cases include: (1) Tension loading on a single-edged notched specimen -- referred to as the ``Tension case''; (2) shear loading on a single-edged notched specimen -- referred to as the ``Shear case''; and (3) uniaxial tension loading on a domain with internal flaws where crack coalescence is expected -- referred to as the ``Crack coalescence case''. We show our results in Fig.~\ref{fig:benchmarks} and provide additional details below.
As briefly mentioned in Section~\ref{subsec:pfm}, an extremely fine mesh is required in the crack transition zones to accurately compute the crack surface energy. The literature suggests that the mesh size in the cracked region should be at least $h \le l_0/2$ as suggested in~\cite{borden2012phase, miehe2010phase}. Another important aspect in implementing PFM models is the definition of initial pre-existing cracks. Consistent with previous reports from the literature~\cite{wu2020phase}, different methods for defining the initial crack can result in a different force displacement response, which may explain small discrepancies between our force-displacement curve results and the results from other implementations. In this work we have used the initial strain history defined in~\cite{borden2012phase}

\begin{equation}
\mathcal{H}_0(x) = \frac{\phi_c}{1-\phi_c} 
\begin{cases} 
\frac{G_c}{2 l_0} \left( 1 - \frac{2d(x)}{l_0} \right) & \text{if } d(x) \leq l_0/2, \\ 
0 & \text{if } d(x) > l_0/2.
\end{cases}
\end{equation}

Here, $\phi_c = 0.999$ represents the phase field scalar value assigned to pre-existing cracks, while $d(x)$ denotes the shortest distance between points within the domain and the crack.

\begin{figure}[htbp]
    \centering
    \includegraphics[width=0.8\linewidth]{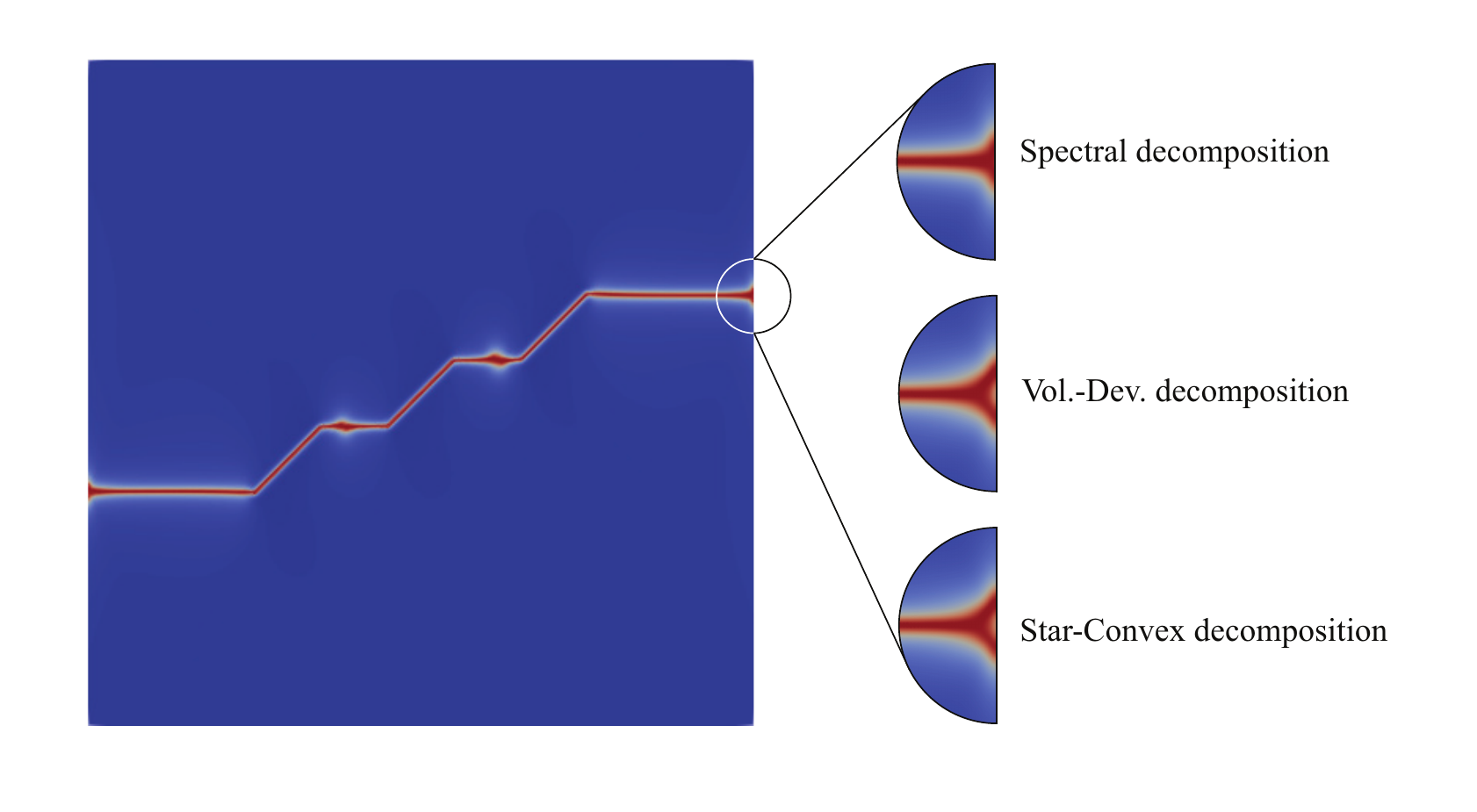}
    \caption{Visualization of crack branching near the boundaries for volumetric-deviatoric and star-convex decomposition. Note the qualitative differences between the three approaches.}
    \label{fig:branching}
\end{figure}
\subsubsection*{Tension case}
The tension case domain is illustrated in Fig.~\ref{fig:benchmarks} a-i. Specifically, it is a $1\times1$ mm square containing an initial crack with length 0.5 mm in the middle of the left edge. The top boundary experiences a uniform vertical displacement with a rate of $\Delta u = 1\times 10^{-5}$ mm for the first 500 steps, and $\Delta u = 1\times 10^{-6}$ mm for the next 1500 steps. For this case, we use an unstructured mesh with linear triangular elements refined in the central band of the domain where the crack is expected to propagate. The smallest element size inside the region is $h=0.001$ mm, and the largest element size outside the crack zone is $h = 0.01$ mm. The convergence criteria tolerance value was set to $1\times 10^{-4}$. The initial crack was modeled by using an initial strain history field following Borden et al.~\cite{borden2012phase}. All other parameters were set to match the recent literature~\cite{ambati2015review} with Young's modulus set to $E = 210$ GPa, $\nu = 0.3$, $G_c = 2.7$ N/mm and characteristic length $l_0 = 4\times 10^{-3}$ mm. As shown in Fig.~\ref{fig:benchmarks} a-ii, where the typical resulting crack pattern is illustrated, and in Fig.~\ref{fig:benchmarks} a-iii where the force displacement curves are illustrated, the results obtained from our simulations for the different energy decompositions are in good agreement with the literature~\cite{ambati2015review}.
\begin{figure}[htbp]
    \centering
    \includegraphics[width=\linewidth]{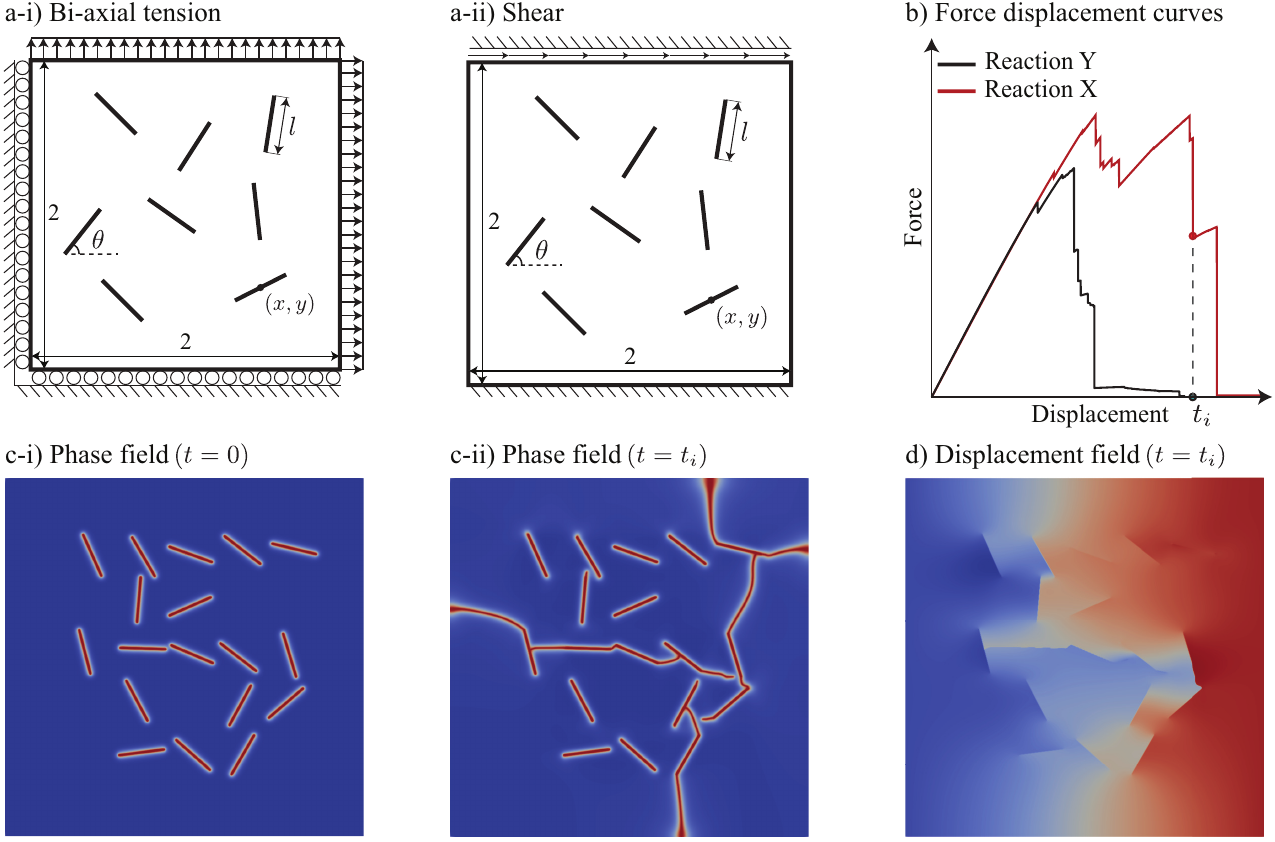}
    \caption{Overview of our dataset: a) Schematic showing the boundary conditions and the geometric properties of the domain, with $n$ number of cracks, where the $\theta$ orientation and center coordinate $(x, y)$ of each crack is randomly sampled for each simulation case. The crack length $l = 0.25$ mm  is fixed in all simulations, b) For each simulation, phase field and x and y components of the displacement in multiple loading steps and the boundary force vs. boundary displacement curves are saved.}
    \label{fig:dataset}
\end{figure}
\subsubsection*{Shear case}
For the Shear case, all physical parameters (Young's modulus $E$, critical energy release rate $G_c$, characteristic length $l_0$ and the poisson's ratio $\nu$) match the tension case. We use a uniform structured mesh with $256\times256$ Q1 elements, following the original implementation by~\cite{ambati2015review}. The top boundary is subject to shear with a constant rate of $\Delta u = 1\times 10^{-5}$ mm. The boundary conditions are shown in Fig.~\ref{fig:benchmarks} b-i , a representative resulting crack pattern is shown in Fig.~\ref{fig:benchmarks} b-ii, and the load displacement curves are shown in Fig.~\ref{fig:benchmarks} b-iii. Consistent with the literature~\cite{ambati2015review}, there are differences in the force displacement curve response using volumetric-deviatoric decomposition when compared to the other models. The spectral decomposition and the star-convex decomposition approaches show a similar response.
\subsubsection*{Crack coalescence case}
Beyond the standard Tension and Shear benchmark cases, we also implemented a benchmark case from Zhou et al.~\cite{zhou2018phase} due to its similarity to the final dataset proposed in this work. In Zhou et al., initial flaws are modeled as geometry-induced cracks inside the mesh, but in our model we use a strain history field for defining the initial cracks. To closely match the force-displacement curve presented in Zhou et al.~\cite{zhou2018phase}, we used a characteristic length of $l_0 = 0.12$ mm and a regionally refined unstructured mesh with triangular elements where the smallest element size within the region is $h=0.05$ mm and the largest mesh element size is $h=0.25$ mm. As illustrated in Fig.~\ref{fig:benchmarks} c-i, the domain is under uniaxial tension where we apply uniform vertical displacement with a rate of $\Delta u = 5\times 10^{-7}$ mm on the top and bottom boundary and fix the x component of the displacement on the bottom boundary. Young's modulus is set to $E = 30$ GPa, $\nu = 0.333$ and $G_c = 3\times 10^{-3}$ N/mm. As can be seen in Fig.~\ref{fig:benchmarks} c-ii-iii, the crack pattern and the force displacement curves are in good agreement with Zhou et al.~\cite{zhou2018phase} which uses the spectral decomposition approach. However, it is worth pointing out that in this case, using either the volumetric-deviatoric or the star-convex decomposition results in branching of the crack pattern near the boundaries as shown in the Fig.~\ref{fig:branching}.
\revs{As a brief note, we ensured that we observed the same branching behavior for these two decompositions using a finer mesh than the one shown in Fig.~\ref{fig:branching} with $h=0.02$ mm and a smaller loading increment of $\Delta u = 1\times 10^{-7}$ mm.}

In our work, all simulations are implemented using the opensource finite element package FEniCSx~\cite{baratta2023dolfinx, scroggs2022construction, scroggs2022basix, alnaes2014unified}.
\begin{figure}[htbp]
    \centering
    \includegraphics[width=1\linewidth]{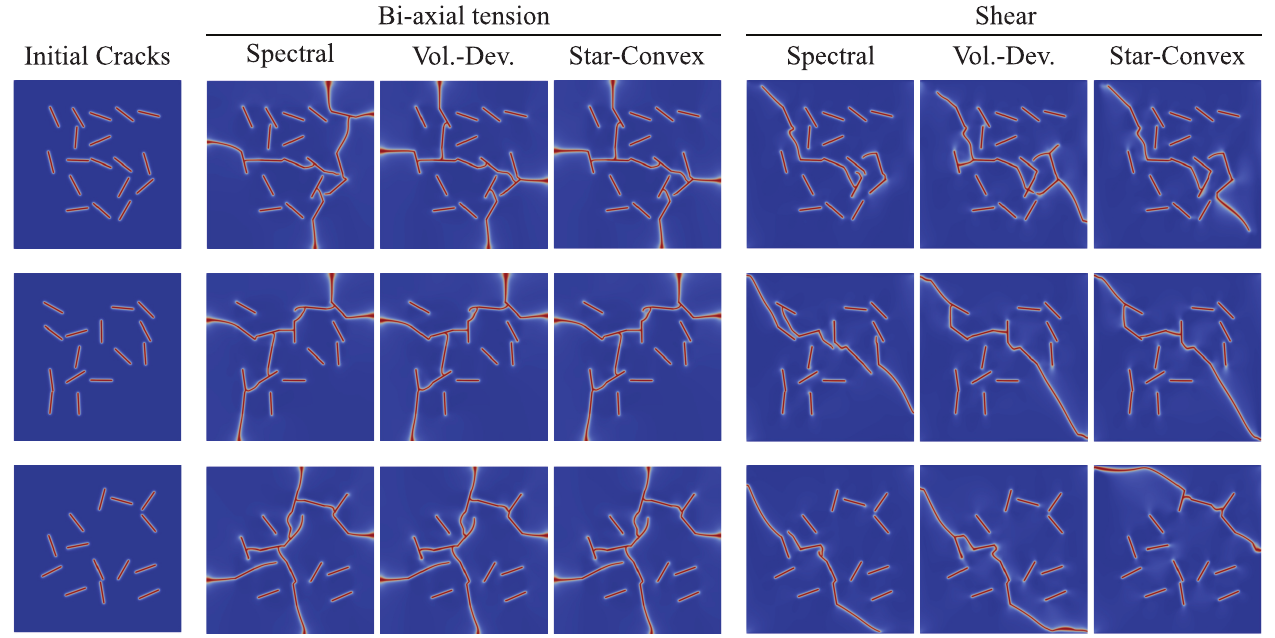}
    \caption{Figure showing the initial crack pattern and final crack pattern for three samples of each dataset subcategory. From these illustrative results we see that in some cases almost no difference between the three energy decomposition methods (see $3^{rd}$ row of the Bi-axial tension case and $2^{nd}$ row of the Shear case), moderate difference ($1^{st}$ row of Bi-axial tension case and Shear case) or extensive difference ($3^{rd}$ row of Shear case) in the resulting final crack pattern.}
    \label{fig:different-cracks}
\end{figure}
\subsection{Dataset specifications}
\label{sec:dataset_spec}
The version of our dataset published in conjunction with this paper contains 6,000 two dimensional quasi static simulations of phase field modeling for brittle fracture. We simulate domains with internal initial cracks under the Bi-axial tension and Shear boundary conditions. For the Shear case, the top and bottom boundaries are also subject to $\phi = 0$ Dirichlet boundary condition as in many examples without this condition, the fracture initiated and propagated along these boundaries, resulting in uninformative fracture patterns that are not useful for training ML models. We run all three energy decomposition methods (1000 initial conditions $\times$ 2 boundary conditions $\times$ 3 energy decomposition methods $=$ 6000 simulations in total). Schematics of representative geometries and applied boundary condition are shown in Fig.~\ref{fig:dataset}a. 
The overall dataset structure is as follows. First, each simulation is categorized under one of 6 sub-categories named as \verb|case-energy_decomp| for example: \verb|tension-spect|. Each sample is saved under the random seed that was used for generating the random internal cracks in order to enhance reproducibility. The random parameters in each sample are the number of internal cracks which is an integer between 10-20, and the coordinates and orientations of the crack inside the domain. 
\revs{All cracks have the same length $l = 0.25$ mm. As a brief note,  this consistent initial crack length was chosen after preliminary investigations in order ensure reliable convergence of the simulations, and avoid situations where crack propagation is dominated by a single large initial crack.}

\revs{For each sample across all datasets, 
we saved 100 snapshots of the crack evolution. All simulations used loading increments of $\Delta u = 1\times 10^{-6}$ mm. The datasets with the Bi-axial tension boundary condition were simulated for 5,000 steps with a saving frequency of 50. The simulations based on the Shear boundary condition were simulated for 10,000 steps with a saving frequency of 100. In total, for all simulated samples, we saved 100 loading steps. For each saved step, we recorded the phase field, and the $x$ and $y$ components of the displacement field. The force displacement data on the support edges were also recorded for all loading steps.} The data is saved in \verb|HDF5| format. The simulation parameters including Young's modulus $E$, Poisson's ratio $\nu$, critical energy release rate $G_c$, characteristic length $l_0$ and the mesh resolution are listed in Table~\ref{tab:sim_params}. A uniform structured mesh with bi-linear elements were used with 4 quadrature points in each element.

\begin{table}[htbp]
    \centering
    \caption{Material parameters and mesh resolution used in the simulations. Please note that the code to reproduce these data is also freely available, see Section~\ref{sec:additional_info}.}
    \begin{tabular}{@{}lc@{}}
        \toprule
        \textbf{Parameter} & \textbf{Value} \\ 
        \midrule
        $E$ (Young's modulus) & 1000 GPa \\ 
        $\nu$ (Poisson's ratio) & 0.3 \\ 
        $G_c$ (Fracture energy) & 1 N/mm \\ 
        $l_0$ (Length scale) & 0.01 mm \\ 
        $\Delta u$ (Loading increment) & $1 \times 10^{-6}$ mm \\
        Mesh resolution & $800 \times 800$ Q1 \\ 
        \bottomrule
    \end{tabular}
    \label{tab:sim_params}
\end{table}

\begin{table}[ht]
    \centering
    \caption{\revs{Simulation specifications for the two boundary conditions and energy decomposition methods. Here, $N_s$ denotes the number of samples in each dataset and $N_t$ is the number of saved snapshots in each trajectory.}}
    \label{tab:sim_specs}
    \resizebox{\textwidth}{!}{%
    \begin{tabular}{c l c c c c c c c c}
        \toprule
        \textbf{Boundary Condition} & 
        \textbf{Decomposition} &
        \textbf{Channels} &
        \textbf{Resolution} &
        \textbf{$N_s$} & 
        \textbf{$N_t$} & 
        \textbf{Avg. Time (h/sim)} & 
        \textbf{Hardware} & 
        \textbf{Dataset Size} \\
        \midrule
        \multirow{3}{*}{\centering Bi-axial Tension} 
            & Spectral     & $\phi, u, v$ & $128 \times 128$ & 1000 & 100 & 2.88 & 16-core CPU & 32.4 GB \\
            & Vol.-Dev.    & $\phi, u, v$ & $128 \times 128$ & 1000 & 100 & 2.76 & 16-core CPU & 32.4 GB \\
            & Star-convex  & $\phi, u, v$ & $128 \times 128$ & 1000 & 100 & 2.74 & 16-core CPU & 32.4 GB \\
        \midrule
        \multirow{3}{*}{\centering Shear} 
            & Spectral     & $\phi, u, v$ & $128 \times 128$ & 1000 & 100 & 5.42 & 16-core CPU & 33.7 GB \\
            & Vol.-Dev.    & $\phi, u, v$ & $128 \times 128$ & 1000 & 100 & 5.26 & 16-core CPU & 33.4 GB \\
            & Star-convex  & $\phi, u, v$ & $128 \times 128$ & 1000 & 100 & 5.28 & 16-core CPU & 33.6 GB \\
        \bottomrule
    \end{tabular}%
    }
\end{table}

The dataset can be accessed in both a $128\times128$ grid format, and an original mesh resolution format. The $128\times128$ format offers a good balance of information and data size maintaining a standard resolution for ML applications and ease of use, while the original mesh resolution format enables flexibility. In Fig.~\ref{fig:different-cracks}, we also show that different energy decompositions sometimes -- but not always -- result in significantly different outcomes. \revs{All the simulations were run in parallel using 16 cores. The parameters regarding the simulation time, hardware used, and dataset size are listed in Table~\ref{tab:sim_specs}}

\subsection{Baseline Models}
\label{sec:baseline}
In this Section, we briefly summarize the baseline models implemented on these data. Following the example set by PDEBench~\cite{takamoto2022pdebench}, we choose three standard baselines: An extension of Physics Informed Neural Networks (PINNs) called Deep Ritz Method (DRM)~\cite{yu2018deep}, Fourier Neural Operators (FNOs), and U-Net convolutional neural networks (UNet). Standard ``out-of-the-box'' implementations of these baseline models are not designed with our phase field fracture dataset in mind. Thus, in all three cases, we elected to iterate on architecture and hyperparameters up to a point. Specifically, our goal was to make a balanced investment in baseline model selection -- sufficient effort to generate meaningful results, while leaving advanced machine learning model development in this area to future work by both us and others. In addition, we made the pragmatic choice of prioritizing model implementations where the code was published alongside the paper. All the models were implemented using PyTorch~\cite{paszke2019pytorch} and the process of logging and hyperparameter tuning was done using the Weights and Biases tool~\cite{wandb}. After preliminary investigation, we selected a recently published PINN formulation designed specifically for phase field fracture problems~\cite{manav2024phase}, the ``out-of-the-box'' implementation of FNOs from PDEBench~\cite{takamoto2022pdebench}, and the UNet architecture used in our previous work on calibration for classification problems in mechanics~\cite{mohammadzadeh2023investigating}. Further details follow.

\subsubsection{Physics Informed Neural Networks}
\label{sec:PINN}
The goal of training a Physics-Informed Neural Network (PINN) is to approximate the solution to a partial differential equation (PDE).
The concept was introduced in~\cite{raissi2019physics}, and since then there has been an explosion of related work in the literature~\cite{zhang2022analyses, rasht2022physics, kharazmi2021hp, cai2021physics, karniadakis2021physics}. In some cases, neural networks have been used to minimize the residual of the strong form of the PDE, as in~\cite{raissi2019physics}, while in others, the variational form of the PDE has been used, as in~\cite{kharazmi2021hp, khodayi2020varnet, yu2018deep}.
There have been multiple recent attempts to apply PINNs to phase field models of brittle fracture, for example~\cite{goswami2020adaptive, goswami2022physics, chakraborty2022variational, manav2024phase, goswami2020transfer, ghaffari2023deep, zheng2022physics}. In some of these studies, to mitigate the computational cost of training a PINN for every loading step, different transfer learning approaches have been explored. For instance, the trained weights from one loading step can be used as a good initialization point for the next step~\cite{manav2024phase}, or the weights of all layers except the last can be kept fixed after training during the first step, and then only the weights of the last layer of the feedforward network are retrained~\cite{goswami2020transfer}.
A typical PINN architecture consists of multilayer perceptrons (MLPs), and usually weighted combinations of multiple loss terms that are mainly physics-informed as a form of a PDE residual and can also have supervised data-driven terms, on a set of collocation points inside the domain or on the boundary~\cite{raissi2019physics}. PINN based models are usually trained for a large number of iterations to drive the residual toward zero. 
\revs{Here we should note that a PINN based model is trained to approximate the solution to a PDE for a specific set of input functions. Operator learning approaches, such as DeepONets~\cite{lu2019deeponet} and Fourier Neural Operators~\cite{li2020fourier}, can utilize physics informed loss terms similar to the PINNs to be able to generalize over a wider range of input functions without the need for retraining.}

For our representative baseline model, we explore the process of minimizing the PINN loss function for only a single set of boundary conditions and initial cracks over a predefined number of loading steps.
In our preliminary explorations (not reported in detail in this paper), we attempted both the vanilla PINN formulation using the DeepXDE Python package~\cite{lu2021deepxde}, and the DRM model developed by Manav et al. ~\cite{manav2024phase}. The DRM uses the quadrature points of the mesh as the collocation points, and is formulated to minimize the energy functional of the variational problem~\cite{goswami2020adaptive, manav2024phase}. For the main PINN baseline model presented in this paper, we have used the PINN architecture and code provided by Manav et al.~\cite{manav2024phase} as a representative example from the recent literature of a PINN formulation with phase field fracture as the target application. We chose this approach because the implementation was well documented and openly available and was benchmarked on the crack coalescence case which aligned closely with our dataset.
\revs{To make a fair comparison, a uniform unstructured mesh with the same mesh size as the FEA solution described in Section~\ref{sec:dataset_spec} with triangular elements was provided as the input to the model. After experimenting with different hyperparameters such as width and depth of the hidden layers, we settled on 6 hidden layers, each having 100 neurons and with characteristic length set to 0.01 in the non-dimensionalized energy functional as described in~\cite{manav2024phase}. As a brief note, training a model with more parameters was not realistic with the computational resources that we have access to. This is consistent with the paper on which this model is based~\cite{manav2024phase}, where the authors noted that training a model can take up to an order of magnitude longer for much simpler crack pattern cases.}

\revs{The trained models use adaptive ReLU as the activation function. Instead of using auto-differentiation, the gradients of the nodal values in each element, approximated by the model, were calculated using linear shape functions. These gradients were then used to approximate the integral within each element via the Gaussian quadrature rule. The L-BFGS~\cite{liu1989limited} optimizer was used for the first step, and the remaining steps used RPROP~\cite{riedmiller1993direct} as the optimizer.}
We note briefly that though this method has demonstrated efficacy on simpler initial crack patterns (e.g., single-edge notched specimen under tension and shear and crack coalescence), it has not been rigorously tested on more complex patters such as the ones provided in our dataset.

\subsubsection{Fourier Neural Operator (FNO)}
\label{sec:fno}
Fourier Neural Operators were proposed for operator learning, mapping infinite-dimensional input functions to output functions~\cite{li2020fourier}.
Fourier Neural Operators (FNOs) are among a plethora of operator learning approaches that parametrize the integral kernel in the Fourier space using neural networks~\cite{kovachki2023neural, lu2021learning, shih2025transformers}. 

In this work, we have followed the PDEBench~\cite{takamoto2022pdebench} implementation of basic unrolled autoregressive training and used the same architecture and hyperparameters as the original paper by Li et al.~\cite{li2020fourier}. Specifically, this implementation has four Fourier layers, with the modes in each direction set to 12 with Gaussian Error Linear Unit (GELU) activation functions. We trained the model for 300 epochs with the mean squared error (MSE) loss and the Adam optimizer~\cite{kingma2014adam} setting the initial learning rate to 0.001 with a scheduler set to halve the learning rate every 100 epochs, and a batch size of 4. We trained the FNO model using unrolled autoregressive training approach with a 10 step history length. The model is trained for the complete trajectory on the training set. We feed all three channels for the input (phase field, displacement field in x direction, and displacement field in y direction) for five different seeds. 
\revs{In addition, for each loading step, we scaled the input and output displacement fields of the FNO to the range [0, 1], using the applied boundary displacement at that step as the reference maximum value.}

\subsubsection{UNet}
\label{sec:unet}
Here we use the UNet architecture implemented in~\cite{mohammadzadeh2023investigating}.
Due to their ease of implementation and popularity in the broader literature~\cite{takamoto2022pdebench, ohana2024well}, UNets are a natural choice as a baseline model~\cite{ronneberger2015u}. After preliminary exploration of multiple potential UNet architectures and loss functions, we settled on the UNet architecture used in previous work by our group for investigating model calibration for classification~\cite{mohammadzadeh2023investigating}. In contrast to our FNO implementation, we did not train the UNets autoregressively because in our initial attempts, unrolled autoregressive training led to unstable behavior. 
And, following preliminary exploration, we only trained the UNets with the phase field channel. Because the UNet struggled with the limited number of training samples, we applied data augmentation such as random rotations and horizontal and vertical flips on the fly to improve its performance. In contrast, the FNOs did not require such augmentation, since they were trained autoregressively over the time steps and did not suffer from the same limitation.  We set the batch size to 16 for 100 epochs using the Adam optimizer with the learning rate set to 0.0001.  For the UNet we used the sum of Dice and Focal loss with $\gamma = 2$\ and $\alpha = 0.8$~\cite{lin2017focal} with equal weights and we used 0.4 threshold for binarizing the ground truth. The same dataset and data split was used for the UNet as the FNO.

\subsection{Ensembling methods}
\label{sec:ensembling-methods}
We trained both the FNO and UNet models with five different random seeds. To showcase performance in this paper, we used the dataset with the Bi-axial tension boundary condition and the spectral decomposition energy split with a $128\times128$ input resolution. The dataset was split into $80\%$ training data, with the remaining $20\%$ split into $60\%$ as the meta train split to train the stacking ensemble, $20\%$ as the meta validation set for tuning the hyperparameters of the stacking model, and the final $20\%$ used as the meta test set. As seen in Fig.~\ref{fig:unet-best}, models trained with different random seeds often predict different results. \revs{Compellingly, and consistent with our previous work in the area~\cite{mohammadzadeh2023investigating}, the multiple different results often look ``plausible'' which motivates the general idea of exploring methods that can combine these different results in a manner that will enable future study from an uncertainty quantification perspective~\cite{lakshminarayanan2017simple}.}
In line with the goals of this paper, we provide baseline results from performing straightforward ensemble averaging across the five model seeds using majority and unweighted averaging also known as hard and soft voting respectively~\cite{lam1995optimal, ganaie2022ensemble} and stacking methods~\cite{wolpert1992stacked}. A brief explanation of each approach follows.

Hard voting is a well known ensembling method also referred to as ``majority voting''. In this simple approach, we aggregate the class label predictions from different models, and the majority vote determines the final prediction. Soft voting is another well known and straightforward ensembling method that averages the class label probabilities across different models~\cite{lam1995optimal}. It has been observed that even using these simple ensembling methods can improve the performance of the models such as UNet~\cite{mohammadzadeh2023investigating}, and different random initializations of the models can lead to learning different patterns and modes of the data and ensembling the results of them would reduce the variance of the final prediction~\cite{fort2019deep, prachaseree2022learning, ganaie2022ensemble}. Finally, stacking, originally introduced in~\cite{wolpert1992stacked}, is an ensembling method that trains a meta-learner on the predictions of the base models to learn a linear or nonlinear combination of their outputs, producing a final prediction for the class label. To perform our baseline model of stacking, we used a simple architecture with 81 parameters that is comprised of two 2D convolutional layer with kernel size 1 and 3 to denoise and capture local features on all five channels and then a ReLU activation function and a final 2D convolutional layer that aggregates all five channels into the final output. We have observed that in this small network the initialization is very important to get a good result. As the probability of having a cracked pixel inside the domain is very low (approximately $2\%$) we initialized the weights of the first and the last layer using Xavier initialization~\cite{glorot2010understanding}. Additionally we initialize the bias of the first layer to 0.2, enabling the model to start from a soft voting state in which each of the five sub models is given equal weights. The $3\times3$ depthwise layer is initialized with the identity stencil with no bias. The reasoning behind this initialization was so that the initial pass through the network would start from a stable point and to prevent dead ReLU activations. We trained the stacker model for 50 epochs on the meta train split of both UNet and FNO predictions. As the stacking method in its most basic form can approximate soft voting by learning equal weights for each channel, it should be noted that its true potential lies in extending it with added contextual windows and nonlinearities. This allows the network to learn a nonlinear combination of the sub models' outputs and to learn from weaknesses and strengths of them across different regions of the field. Though many more advanced ensembling techniques exist~\cite{qin2024toward, sharma2024ensemble, huang2017snapshot} and are considered ``state of the art,'' we consider them beyond the scope of this paper where our goal is to provide a set of baseline models.\revs{In Table~\ref{tab:baseline_models}, we summarize the characteristics of the three baseline models and their training schemes.}

\begin{table}[ht]
    \centering
    \caption{\revs{Characteristics of the baseline models and training schemes used for comparison.}}
    \label{tab:baseline_models}
    \resizebox{\textwidth}{!}{%
    \begin{tabular}{lccccccc}
        \toprule
        \textbf{Model} & \textbf{Parameters} & \textbf{Input} & \textbf{Output} & \textbf{Activation} & \textbf{Training scheme} & \textbf{Loss} & \textbf{Reference} \\ 
        \midrule
        DRM PINN & 51,109 & Unstructured triangular mesh & $\phi, u, v$ & Adaptive ReLU & Transfer learning & Log residual & \cite{manav2024phase} \\
        FNO      & 465,557 & $\phi, u, v$ & $\phi, u, v$ & GELU & Autoregressive & MSE & \cite{li2020fourier} \\
        UNet     & 1,926,689 & $\phi$ & $\phi$ & ReLU & One-shot & Dice + Focal & \cite{ronneberger2015u} \\
        \bottomrule
    \end{tabular}%
    }
\end{table}

\begin{figure} [h]
    \centering
    \includegraphics[width=\linewidth]{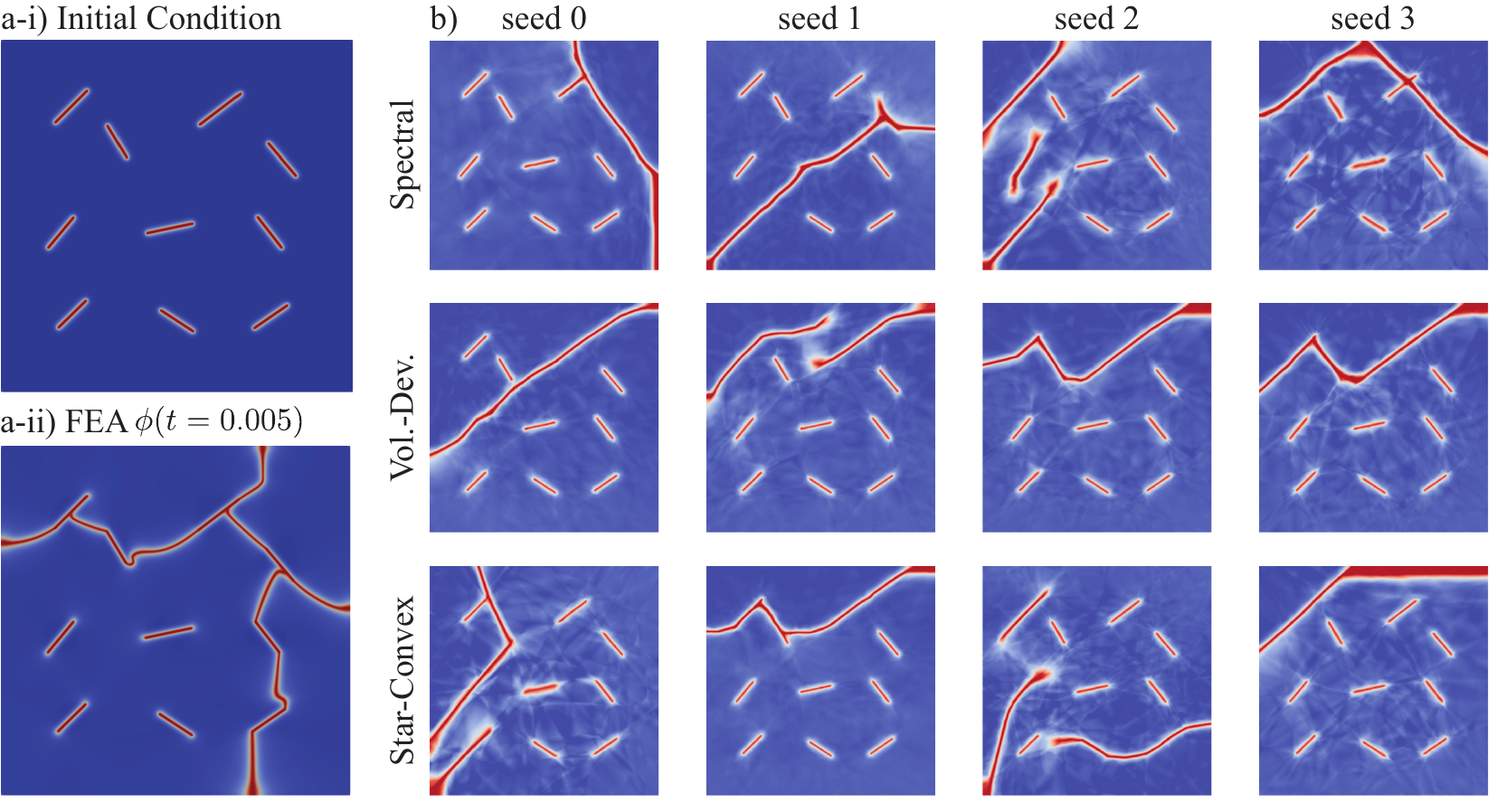}
    \caption{Figure showing: a) The initial condition and the resulting FEA simulation under the Bi-axial tensile loading and b) The results of training the PINN model (12 separate models) with three energy decomposition methods across four different seeds. We note that the individual PINN models are converging to different crack patterns. Though some of these models show plausible results, none were able to predict the correct crack pattern.}
    \label{fig:pinn-best}
\end{figure}

\section{Results and Discussion}
\label{sec:results}
In this Section we investigate the results of training the baseline models and discuss approaches to improve and evaluate their performance. Sections~\ref{sec:pinn-result} to~\ref{sec:unet-result} present the performance, limitation and potential future directions for PINNs, FNOs and UNets. Then in Section~\ref{sec:ensemble-result}, we investigate the improvement as a result of using different ensembling approaches. In Section~\ref{sec:comparisons}, we provide a comparison of the baseline models, and finally in Section~\ref{sec:metrics} we briefly go over the implications of using different error metrics for ML model training and evaluation.
\begin{figure} [p]
    \centering
    \includegraphics[width=\linewidth]{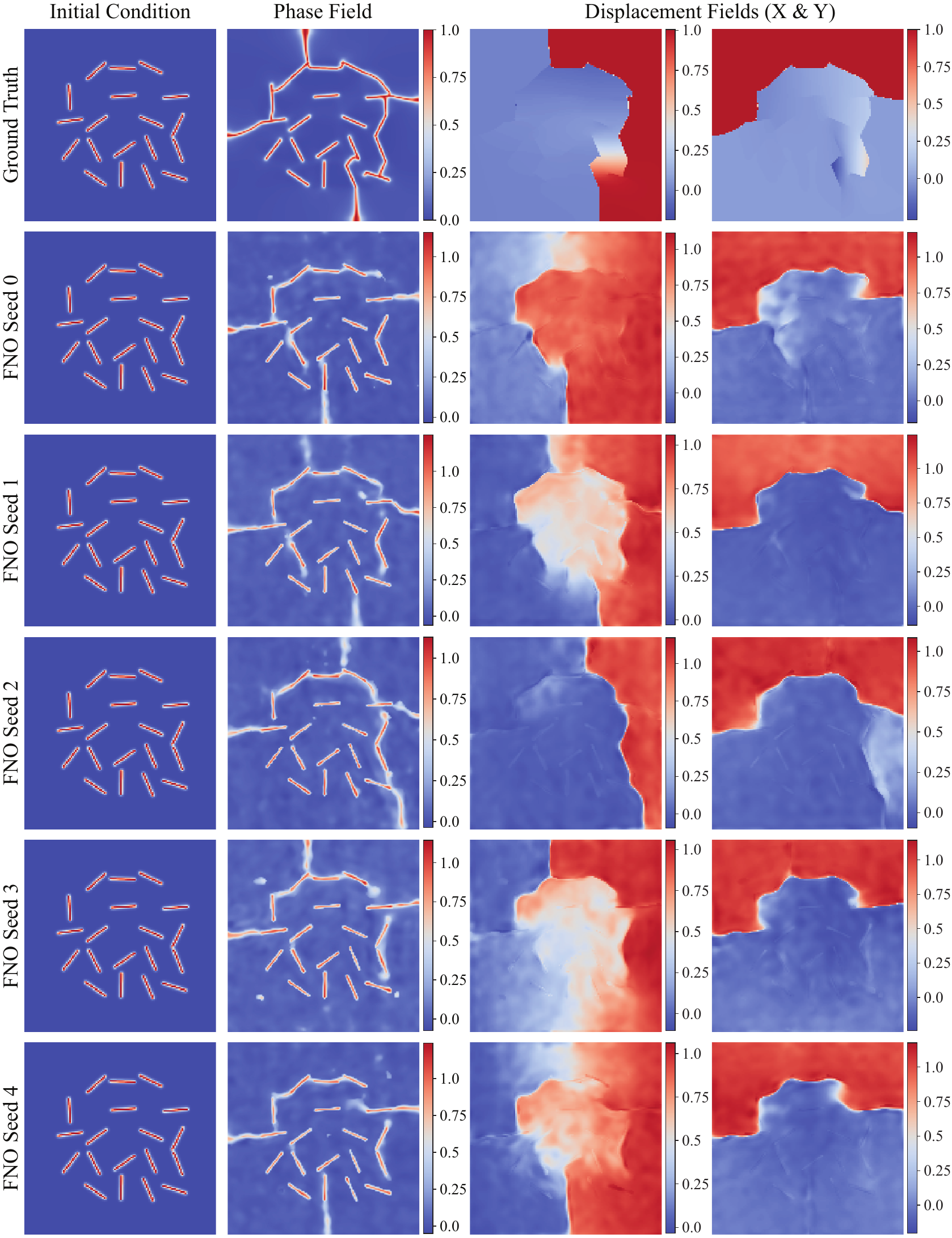}
    \caption{This figure presents the predicted phase field and displacement fields for each FNO model that was trained on the dataset. \revs{This shows the results for a representative sample at the $50^{th}$ percentile of performance with a Dice score of 0.6869.} We can see better performing samples in Fig.~\ref{fig:appendix-fields-1} The top row shows the ground truth and each subsequent row shows the results of each FNO seed. Additional examples are shown in Fig.~\ref{fig:appendix-fields-1}}
    \label{fig:fno-raw}
\end{figure}
\subsection{Physics Informed Neural Network Baseline}
\label{sec:pinn-result}
To investigate the ability of PINNs to predict phase field fracture behavior, we began with a vanilla implementation using the DeepXDE package~\cite{lu2021deepxde}. Consistent with previous studies~\cite{manav2024phase, ghaffari2023deep}, we found that using the strong form of the PDE only produces meaningful phase field fracture results in the 1D setting. Thus, for the baseline PINN model shown here, we adopted the Deep Ritz method (DRM), proposed by Manav et al.~\cite{manav2024phase}. Specifically, we separately trained 12 models under the Bi-axial tension boundary condition. We investigated all three energy decomposition methods (spectral, volumetric-deviatoric, and star-convex) with four different seeds each. In the FEA setting, these simulations were displacement controlled, and all three energy decomposition strategies resulted in the same crack pattern. In Fig.~\ref{fig:pinn-best}a, both the simulation initial condition and the ground truth FEA output (identical for all energy decomposition methods) are shown.

The results of training the DRM model for all four seeds are shown in Fig.~\ref{fig:pinn-best}b. As can be seen from the illustrations, none of the models converged to the correct solution, and each of them resulted in a distinctly different crack pattern. In the original paper proposing the DRM model~\cite{manav2024phase}, there was a crack coalescence test that showed good agreement with the ground truth finite element solution. However, we note that: (1) the reported PINN crack pattern differs slightly from the finite element derived ground truth, and (2) it is a test with only three initial cracks and is therefore much simpler than the example investigated in Fig.~\ref{fig:pinn-best}. With only three initial cracks (all oriented in the same direction and with even spacing, see Fig.17 in~\cite{manav2024phase})the inaccuracies in approximating the energy landscape are less probed than in the example shown in our paper where errors manifest more clearly in the predicted crack paths.

We note briefly that in designing this study we chose a uniform mesh with the same mesh size as the FEA simulations to train the PINNs ($0.0025$ mm). Due to the high computational cost of training each PINN model (approximately 24 hours to simulate up to the same loading step as the FEA model on one NVIDIA V100 GPU), we did not iterate significantly to tune the network architecture or the number of trainable parameters of the model. 
Critically, we also note that poorly performing examples may still lead to visually convincing but physically incorrect results. This highlights the importance of incorporating additional physics-based constraints and error evaluation approaches, as these misleading results during test-time inference can give a false sense of accuracy and reliability.
Exploring such modifications is one promising direction for improving PINN performance in the future, and we hope that our dataset will encourage further development in the area.
 
\begin{figure}[htbp]
    \centering
    \includegraphics[width=1 \linewidth]{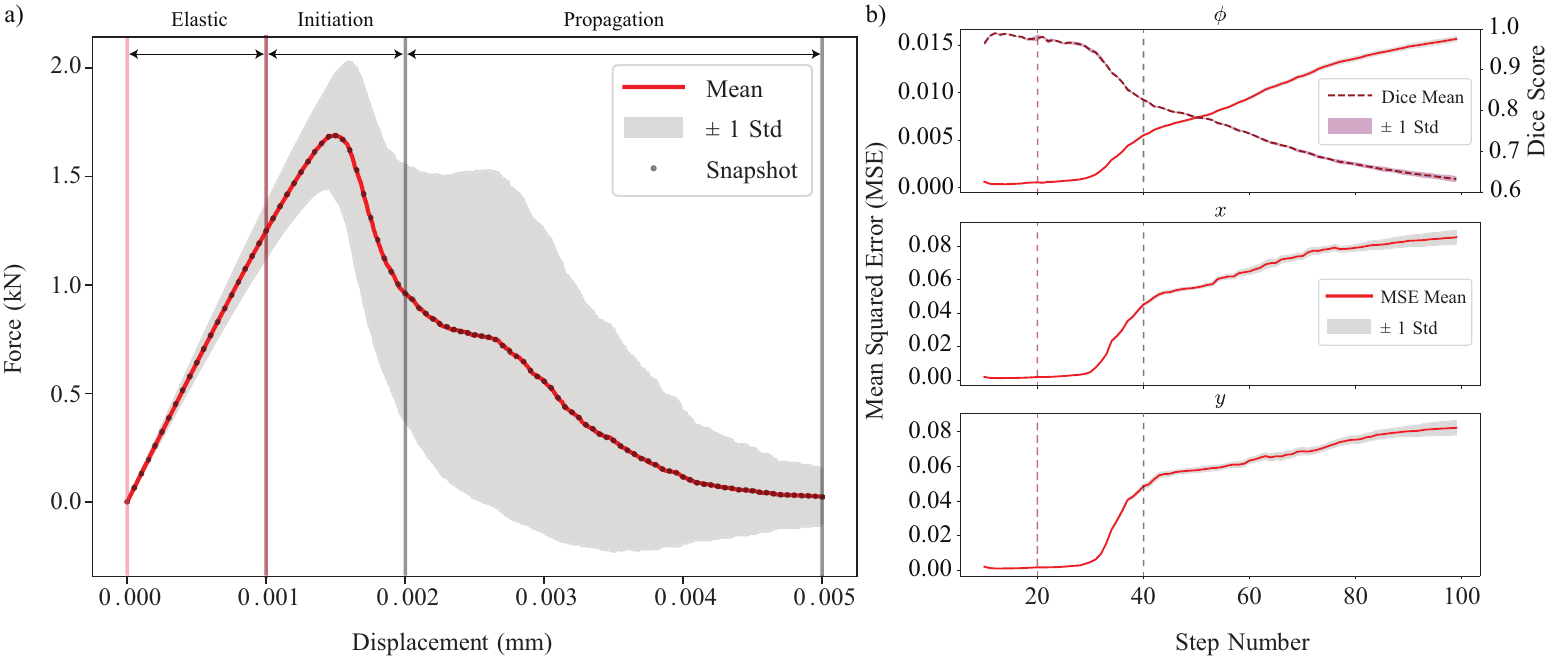}
    \caption{\revs{This figure presents: a) The mean and standard deviation of the force displacement curves across all samples in the dataset, along with the three regions (elastic, initiation, propagation) indicated and b) The rollout error metrics for the FNO models, averaged across the test set and an ensemble of five models, plotted against the simulation step number. The vertical dashed lines indicate the transition between the elastic, crack initiation, and crack propagation regimes. The top panel is a dual-axis plot for the phase field($\phi$), showing the Mean Squared Error (MSE) on the left axis and the Dice score on the right. The middle and bottom panels show the MSE for the $x$ and $y$ components of the displacement field, respectively, illustrating how predictive error remains low during the elastic phase but increases significantly once cracks initiate and propagate. The rollout MSE is computed from a single, continuous prediction of the entire simulation trajectory. To start this process, each model in the ensemble is given the initial 10 steps of the simulation from the ground truth and then predicts the remaining 90 steps autoregressively. For analysis purposes only, the resulting error is then averaged within the three distinct windows shown to illustrate how error accumulates as the simulation progresses from simple to more complex behavior. The specific MSE values for each window are provided in Table~\ref{tab:mse}.}}
    \label{fig:fd-region}
\end{figure}

\begin{figure} [p]
    \centering
    \includegraphics[width=0.9\linewidth]{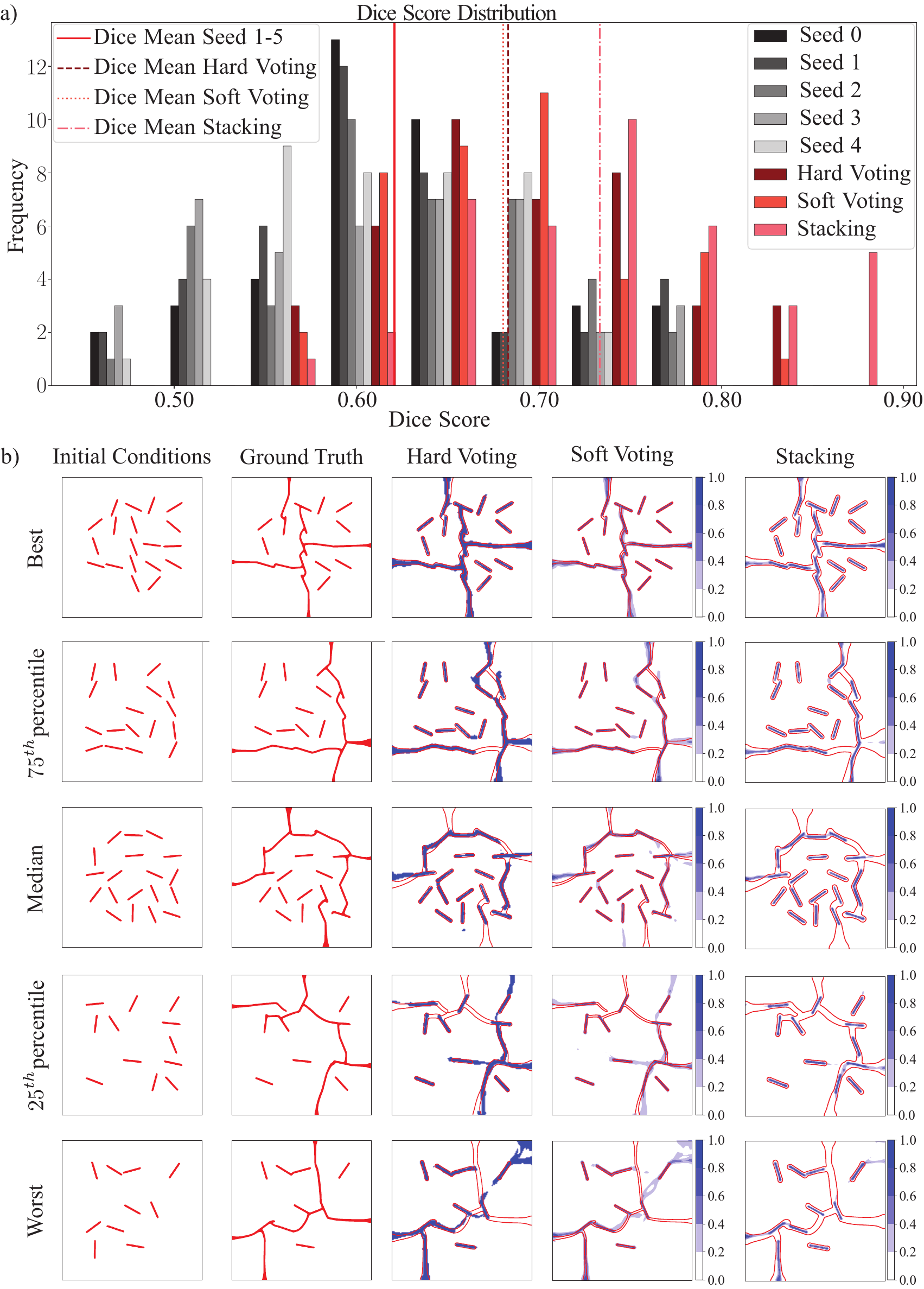}
    \caption{This figure presents: a) The histogram of the Dice score distribution of the FNO models trained with five different seeds and ensembling methods, and b) representative test samples showing the performance of each ensembling method. For each sample, the initial condition is shown in the first column and the corresponding ground truth contour is shown on the second column. The subsequent columns display the result of ensembling methods with the ground truth contour overlaid in red for visual comparison.}
    \label{fig:fno-best}
\end{figure}

\begin{figure}[p]
    \centering
    \includegraphics[width=0.9\linewidth]{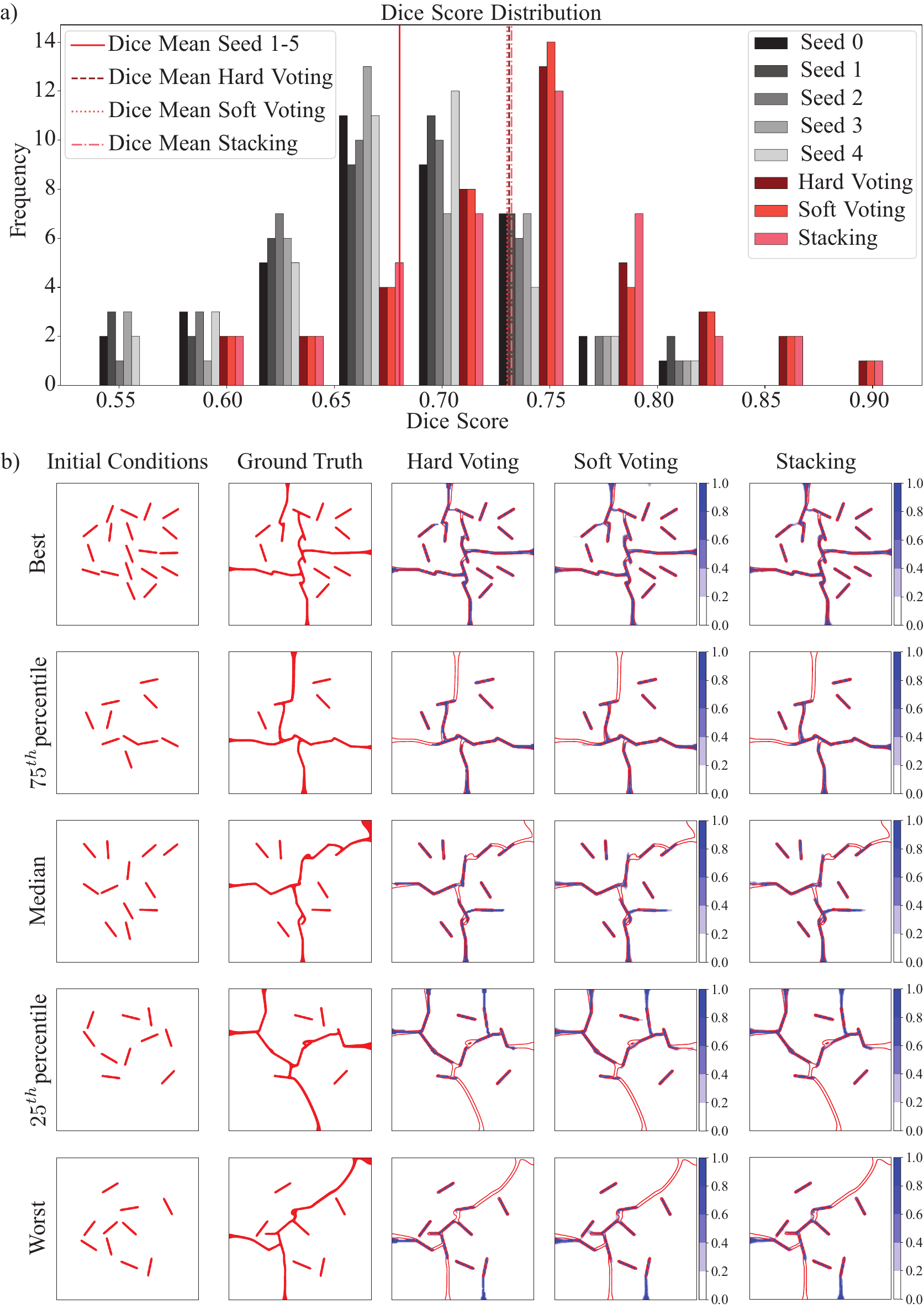}
    \caption{This figure presents: a) The histogram of the Dice score distribution of the UNet models trained with five different seeds and ensembling methods, and b) representative test samples showing the performance of each ensembling method. For each sample, the initial condition is shown in the first column and the corresponding ground truth contour is shown on the second column. The subsequent columns display the result of ensembling methods with the ground truth contour overlaid in red for visual comparison.}
    \label{fig:unet-best}
\end{figure}

\subsection{Fourier Neural Operator Baseline}
\label{sec:fno-result}
Unlike the PINNs approach described in Section~\ref{sec:pinn-result}, we were able to achieve reasonable baseline model results with the standard FNO model described in Section~\ref{sec:fno}. Thus, in this Section, we present both quantitative metrics and qualitative outputs for trained FNO model performance. In Fig.~\ref{fig:fno-best}a-b, we show the quantitative and qualitative performance of the FNO model for all five individual seeds, and the ensemble results across all five seeds. The ensemble results will be discussed in more detail in Section~\ref{sec:ensemble-result}. 

In Fig.~\ref{fig:fno-best}a, we plot a histogram of the Dice score across all 40 test samples evaluated with the trained FNO models. 
As indicated by the vertical line labeled ``Dice Mean Seed 1-5'' in the Fig.~\ref{fig:fno-best}a histogram, the average Dice score across the test set and across all seeds is 0.6208. 
\revs{To further investigate the performance of autoregressive models on this dataset, we divided the simulation trajectory into 3 windows, as shown in Fig~\ref{fig:fd-region}-a. The first 20 steps corresponds to the elastic regime with no crack initiation. The second window (steps 21-40) captures the crack initiation, while the final window (steps 41-100) represents crack propagation within the domain.
        In Fig~\ref{fig:fd-region}-b, we can see that the MSE of all three fields ($\phi, u_x, u_y$) remains low during the elastic regime, but it increases significantly during the crack initiation phase and continues to rise during crack propagation. This is expected, as there is zero to no change in the phase field and very small variance in the displacement field during the elastic regime, making it easier for the model to approximate the solution with low error. However, as the cracks start to initiate and propagate, the phase field and the displacement field undergo significant changes, making it more challenging for the model to predict the crack evolution accurately.
        Figures~\ref{fig:mse-diff-1}-\ref{fig:mse-diff-3} in the appendix provide further insight by illustrating the ground truth at the 100th step alongside the predicted crack patterns from different  FNO seeds in each row and the absolute difference between these two for each channel. These results highlight both the sensitivity of FNOs to initialization, and their difficulty in capturing high frequency features such as the edges of the crack patterns.}
We can see in Fig.~\ref{fig:fno-raw} that FNO models predicted a visually less well-defined and smoother phase field solution when compared to the FEA Ground Truth. A plausible explanation for this outcome is related to a well-known limitation of neural networks referred to as spectral bias, which indicates the preference of neural networks to learn low frequency features at the expense of high frequency features~\cite{rahaman2019spectral, xu2019frequency, xu2025understanding, khodakarami2025mitigating}. Another interesting observation from Fig.~\ref{fig:fno-raw} is that although the predicted displacement fields may deviate from the ground truth, they remain consistent with their corresponding phase fields. This suggests that the models have captured some underlying physical relationships through training on this dataset. 

\begin{table}[h!]
            \centering
            \caption{\revs{Mean Squared Error (MSE) of FNO predictions averaged over trajectory windows and averaged over seeds, reported for the elastic, initiation, and propagation windows.}}
            \begin{tabular}{lccc}
            \toprule
            \textbf{Data Field} & \textbf{Elastic Window (10:20)} & \textbf{Initiation Window (21:40)} & \textbf{Propagation Window (41:100)} \\
            \midrule
            Phase Field      & 7.40e-04 & 5.08e-03 & 1.23e-02 \\ 
            X Displacement   & 2.54e-03 & 4.24e-02 & 7.13e-02 \\
            Y Displacement   & 2.65e-03 & 4.00e-02 & 7.50e-02 \\
            \bottomrule
            \end{tabular}
            \label{tab:mse}
\end{table}

Looking ahead, there are several promising directions to improve upon these baseline models. These include the exploration of various data augmentation strategies, ranging from traditional techniques to more advanced generative AI based approaches. As shown in Fig.~\ref{fig:fno-best}b, which presents crack pattern predictions ranked from best to worst based on the soft voting method (discussed further in Section~\ref{sec:ensembling-methods}), we observe that examples with lower Dice scores typically have large quantitative and qualitative deviations from the ground truth. Thus, although the top-performing predictions (e.g., those in the upper row of Fig.~\ref{fig:fno-best}b) show strong agreement with the FEA ground truth, the substantial discrepancies in lower-performing cases highlight significant room for improvement.

\subsection{UNet Baseline}
\label{sec:unet-result}
In this Section, we evaluate the performance of the UNet models on the same dataset to assess predictive accuracy across different random seeds and ensembling strategies. As outlined in Section~\ref{sec:unet}, the standard UNet architecture without any hyperparameter tuning was trained on the Bi-axial tension dataset with spectral energy decomposition, with five different seeds.

Fig.~\ref{fig:unet-best}a shows the histogram of the Dice score distribution across all 40 test samples evaluated with the trained UNet models and ensembling approaches (discussed in Section~\ref{sec:ensembling-methods}). The UNet models on average scored 0.6804 in Dice score among different seeds. Fig.~\ref{fig:unet-best} also highlights that the models -- likely by design as they are trained on the binary phase field -- produced sharper predictions. Note that unlike the FNO models, the displacement fields were not used during UNet training. 
\begin{figure} [htbp]
    \centering
    \includegraphics[width=\linewidth]{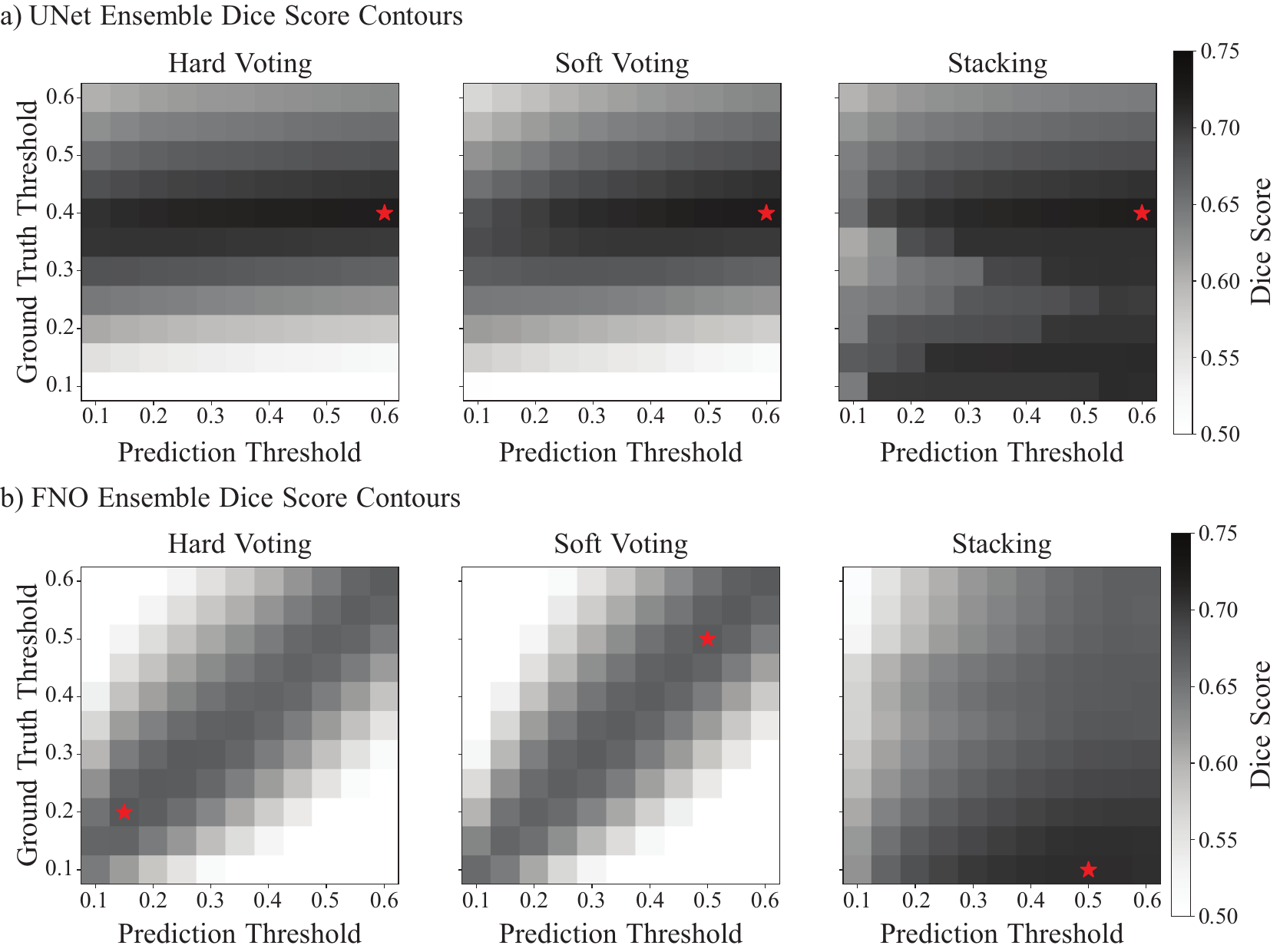}
    \caption{The Figure shows a heatmap of the performance of each ensembling method on the predictions of a) UNet and b) FNO submodels based on the chosen threshold that was used to binarize the prediction and the corresponding ground truth to calculate the Dice score.}
    \label{fig:heat}
\end{figure}
\subsection{Ensemble approaches}
\label{sec:ensemble-result}
In addition to evaluating individual models, we investigated the impact of the three ensembling methods described in Section~\ref{sec:ensembling-methods} (hard voting, soft voting, and stacking) on overall prediction performance.
To compute the Dice score, which requires a binarized prediction field, we were required to threshold both the model prediction fields and the ground truth fields to delineate between ``undamaged'' (i.e., less than threshold) and ``damaged'' (i.e., more than threshold) regions. Because test performance is dependent on this choice of threshold, we performed a threshold search over the range [0.1, 0.6] using the validation set for both the FNO and UNet models separately. The reason for choosing this range was because the models, particularly the FNOs, tended to produce values in range 0.1 to 0.6, as illustrated in Fig.~\ref{fig:fno-raw}. Thus, applying a higher threshold (e.g. close to 0.9) would result in omitting most of the predicted crack regions, as only a small fraction of the predictions would surpass the cutoff. This would lead to very low Dice scores, and would prohibit us from being able to effectively study and compare the performance of the baseline models.
In Fig.~\ref{fig:heat}a and b, we show the resulting heatmaps for selecting both a ground truth threshold and a prediction threshold for the purpose of scoring each ensembling method.
In these plots, the red $\star$ indicates the best performing threshold combination on the validation set, which we selected and then subsequently evaluated on the test set.

\begin{table}[h!]
\centering
\caption{Dice scores of ensembling methods for FNO and UNet models.}
\begin{tabular}{lcc}
\toprule
\textbf{Ensembling Method} & \textbf{FNO} & \textbf{UNet} \\
\midrule
Hard Voting & 0.6831 & 0.7321 \\
Soft Voting & 0.6804 & 0.7315 \\
Stacking & 0.7333 & 0.7325 \\
\midrule
Mean Seeds 1-5 & 0.6208 & 0.6804 \\
\bottomrule
\end{tabular}
\label{tab:Dice}
\end{table}

All ensemble results shown in Fig.~\ref{fig:fno-best}b and Fig.~\ref{fig:unet-best}b and listed in Table~\ref{tab:Dice} reflect these threshold combinations. 
As widely observed in the literature~\cite{prachaseree2022learning, mohammadzadeh2023investigating}, ensembling improves the prediction performance of both model types. However, the magnitude of this improvement differs between the two. For the FNO predictions, Hard voting improved the performance of the model by $\approx 10\%$ to 0.6831, the soft voting by $\approx 9.6\%$ to 0.6804 and the stacking achieved the highest boost, improving the performance by $\approx 18.1\%$ to 0.7333. The stacking method with non-linearity through ReLU activation function and context aware filters as described in Section~\ref{sec:ensembling-methods} outperformed the other ensembling methods. One possible explanation for this outcome is that training a meta learner like the stacker model allows the network to act as a localized router learning which base model's output to prioritize for different types of crack patterns. For example, one initial model might perform better on horizontal cracks while another may more accurately capture crack coalescence. In the future, we could also integrate this stacking model into the architecture itself to form a simple Mixture of Experts (MoE) model~\cite{shazeer2017outrageously}. Looking at both Figs.~\ref{fig:fno-best} and~\ref{fig:unet-best}, we can see that the contextual filters of the stacking model helped to filter out False positives and thus considerably improved the prediction accuracy. 

In contrast, all three ensembling methods resulted in almost identical performance improvement when applied to the UNet. Hard voting improved the Dice score by $\approx 7.5\%$ to 0.7321, soft voting by $\approx 7.3\%$ to 0.7315 and stacking by $\approx 7.6\%$ to 0.7325. We anticipate that this difference is likely due to the lower number of false positives of the individual UNet predictions.

\subsection{Overall Baseline Model Performance -- Methods Comparison}
\label{sec:comparisons}
Standard PINNs using the strong form of the PDE as the loss function only work on the 1D settings of phase field problems. Using the Deep Ritz Method (DRM) to incorporate the energy functional allows training in 2D. but it still fails to achieve accurate solutions in complex cases such as the samples in the current dataset. Moreover, the computational cost of this approach is extremely high (an order of magnitude more than FEA solutions)  making it impractical for many applications.

When comparing the performance of the baseline FNOs to the baseline UNet, we can see that the FNOs produced more smoothed out predictions and struggled to accurately model high frequency features of the domain such as the crack paths.
\revs{This spectral bias is a well known limitation of the FNO architecture~\cite{qin2024toward}. However, in our experiments, the models were still able to capture certain high frequency features in both the phase field and displacement fields as illustrated in the top row of Fig.~\ref{fig:appendix-fields-1}.}
In contrast, the UNet models with Focal-Dice loss function, generally achieved higher Dice scores on the dataset compared to the FNO models and produced sharper predictions.
\revs{The Dice score achieved by the UNet models without ensembling (0.6804) shows a moderately good overlap between the predicted crack patterns and the ground truth. Given the sparsity of the cracked regions, this score represents a non-trivial result and confirms that the UNet models are learning a representation of the underlying pattern, though considerable room for improvement remains.}
The ensembling methods resulted in a larger relative improvement on the FNO predictions of approximately $18\%$ when using the stacking method achieving a comparable performance to the UNet models with a Dice score of 0.7333. This highlights the potential of ensembling methods to mitigate some of the limitations of operator learning architectures especially in challenging prediction tasks.

That being said, there are several promising directions that could be explored to improve upon the baseline model performance and reliability from both the FNO and UNet starting points presented here. First, adopting more advanced UNet-based architectures such as ConvNet UNets~\cite{liu2022convnet} or Multires-WNet~\cite{mohammadzadeh2022predicting}, which extract multi-scale features, could lead to higher performance on these data. Another promising avenue is to incorporate more robust autoregressive training strategies, such as push forward and temporal bundling techniques introduced by Brandstetter et al.~\cite{brandstetter2022message}, to stabilize training over time sequences. Lastly, increasing the number of retained Fourier modes in the FNO architecture may further improve its capacity to model high frequency features and help improve baseline performance.
The purpose of this paper is to showcase a challenging problem, and provide a starting point upon which future improvements can be made. 

\subsection{A Note on Performance Metrics}
\label{sec:metrics}
In addition to potential performance enhancements via iterating on the ML modeling approach, we also note that for the phase field fracture problem specifically there are multiple directions for improvement beyond model selection. For example, in this study we can readily see the inherent limitations of typical performance metrics such as Mean Squared Error (MSE) -- both as the loss function for the network, and as the overall performance evaluation metric. Critically, it has been observed in other studies that MSE tends to emphasize errors of the low frequency features, and encourage learning an averaging approach in the prediction fields~\cite{ledig2017photo, mohammadzadeh2022predicting}. Different approaches have been taken to address the problem of retrieving the high frequency features of the data. These approaches can vary from using multi scale~\cite{cai2019multi, khodakarami2025mitigating} and multi resolution architectures~\cite{mohammadzadeh2022predicting}, using Fourier based metrics~\cite{takamoto2022pdebench}, or Fourier based features~\cite{tancik2020fourier}. 
 
\revs{The limitations of the MSE are particularly pronounced in the phase field channel of the dataset, where the number of cracked pixels is very small compared to intact regions, resulting in highly imbalanced data. To better evaluate the performance of the crack pattern prediction, we use the Sørensen–Dice score, also known as the $F_1$ score in the literature. We note that the Dice score cannot be used effectively on the continuous values such as those found in the displacement field, but can be applied to the binary classification of each pixel in the predicted phase field channel as either damaged or undamaged.} 
This metric is used very commonly in semantic segmentation and classification tasks. This score can be calculated as follows:
\begin{equation}
    DSC = \frac{2|\hat{Y} \cap Y|}{|\hat{Y}| + |Y|} = \frac{TP}{TP + \frac{1}{2}(FP + FN)}
\label{eq:Dice}
\end{equation}
where the $\hat{Y}$ is the binarized prediction field and $Y$ is the binarized target field. Here we can also see the metric in the context of binary classification where TP stands for True Positive, FP stands for False Positive and FN stands for False Negative. The main motivation behind choosing this metric is that the crack patterns occupy a very small portion of the field (approximately $2\%$ of the pixels), with the majority of pixels belonging to the background. This creates a highly imbalanced dataset, making this metric well suited for evaluating performance in fields with sparse, intricate patterns where the true negative classes, the undamaged regions, occupy a large portion of the domain. As can be seen in equation~\ref{eq:Dice}, the Dice score is not affected by the true negatives. While MSE might be a suitable choice for smooth, dense and continuous by nature fields such as displacement fields providing a reliable measure of overall field accuracy, it tends to be dominated by undamaged or background regions of the domain in the phase field channel of the data, making it insensitive to small features of the domain. 

While Dice score provides a useful measure for this case, it has notable limitations. It does not penalize the discontinuous crack paths, for example in Fig.~\ref{fig:fno-best} b all of the samples other than the best performing one show visible discontinuities in the crack trajectory, and also can produce higher scores for samples with more initial cracks or those where the cracks travel a longer and more complex trajectory within the domain. This behavior can be seen in Fig.~\ref{fig:fno-best}b when comparing the best and worst examples. Samples that have more initial cracks tend to get a higher score in comparison to those with a smaller number of initial cracks. These limitations highlight the need for more physically grounded or geometry aware evaluation metrics that account for topological correctness, connectivity, and adherence to physics based constraints. For example, future studies could explore alternative hybrid metrics that combine pixel overlap with physics informed loss components to better assess prediction quality in fracture simulations.

\section{Conclusion}
\label{sec:conclusion}
In this work, we implemented a hybrid phase field model for brittle fracture incorporating three different energy decomposition methods: spectral, volumetric-deviatoric, and star-convex. The model was verified against standard benchmark cases commonly used in the literature and implemented using the open source finite element software FEniCSx~\cite{baratta2023dolfinx, scroggs2022basix, scroggs2022construction, alnaes2014unified}. Based on the verified implementation, we generated six datasets combining two boundary conditions, Bi-axial tension and Shear with the three aforementioned energy decompositions. We then trained three baseline models, PINN, FNO and UNet on these datasets to benchmark their performance. Our results indicate that PINN based approaches have limited applicability in this context, while UNet and FNO models are capable of producing more reasonable predictions. We also study three ensembling methods from simple to more complicated: hard voting, soft voting and stacking. We observe that ensembling methods can improve the performance of the models in general but their performance can differ based on the performance of each submodel. Finally we note the importance, and the effects of metrics used for evaluating and training these models and how they should be used to effectively be able to train and evaluate the performance of ML based solutions. We have published the dataset and open sourced the code to reproduce the dataset and the benchmark models to enable other research groups to build up on them to exceed the baseline performance.

Despite achieving reasonable results with two of the baseline models, we observed that these models are highly sensitive to initialization and are prone to converging to different local minima. This behavior reflects the inherent non-convexity of the underlying optimization problem and suggests that our dataset can serve as a challenging benchmark for evaluating the robustness of different modeling approaches. In particular, it provides a valuable opportunity to study the effects of training hyperparameters, random seeds, loss functions and training strategies on model performance and convergence.
\revs{We anticipate several promising directions for future research, including the design of more data efficient models through tighter integration of governing equations~\cite{hansen2023learning}, developing foundational models for physics~\cite{bodnar2025foundation, mccabe2023multiple, wiesner2025towards} and creating more effective ensembling strategies~\cite{qin2024toward}. Further work could involve designing better metrics and loss functions to mitigate spectral bias\cite{wang2022and}, and training strategies to achieve a more stable rollout for autoregressive models~\cite{brandstetter2022message, lippe2023pde, huang2025physicscorrect}. We also encourage the expansion of the dataset to more complex 2D and 3D fracture scenarios and have open-sourced our code to facilitate this effort.}

\section{Additional Information}
\label{sec:additional_info}
The full resolution and the 128x128 downsampled version of the PFM-Fracture dataset are publicly available on the Harvard dataverse at (\url{https://dataverse.harvard.edu/dataverse/PFM-Fracture}). Due to the large file size, the full resolution dataset is divided into two parts. For example the Shear case with spectral energy decomposition is stored as \verb|shear-spect-fullres-part-1| and \verb|shear-spect-fullres-part-2|. The corresponding downsampled version is named \verb|shear-spect-128|. All code for reproducing the dataset is available on GitHub at (\url{https://github.com/erfanhamdi/pfm_dataset}) and the implementation of baseline models and ensembling methods can be found at (\url{https://github.com/erfanhamdi/pfm_bench}).

\section{Declaration of competing interest}
The authors declare that they have no known competing financial interests or personal relationships that could have
appeared to influence the work reported in this paper.

\section{Acknowledgments}
This work was made possible with funding through the Boston University David R. Dalton Career Development
Professorship, the Hariri Institute Junior Faculty Fellowship, the Haythornthwaite Foundation Research Initiation
Grant, and the U.S. National Science Foundation Cyberinfrastructure for Sustained Scientific Innovation program, under the "Elements: Curating and Disseminating Solid Mechanics Based Benchmark Datasets" project CMMI-2310771 and the U.S. Department of Energy, Advanced Scientific Computing Research program, under the "Resolution-invariant deep learning for accelerated propagation of epistemic and aleatory uncertainty in multi-scale energy storage systems, and beyond" project (Project No. 81824). This support is gratefully acknowledged. We also
acknowledge the support of Saeed Mohammadzadeh for his insightful comments on phase field fracture modeling with FEniCS and data curation, and Boston University's Research Computing Services for providing computing resources.
\newpage
\appendix
\section{Mesh Refinement Study}
\label{appendix:mesh-refinement}
In order to study the effect of the mesh size on the results of the PFM simulations, we perform a mesh refinement study on a single initial pattern for both boundary conditions and all three energy decomposition methods.

\begin{figure}[htbp]
    \centering
    \includegraphics[width=\linewidth]{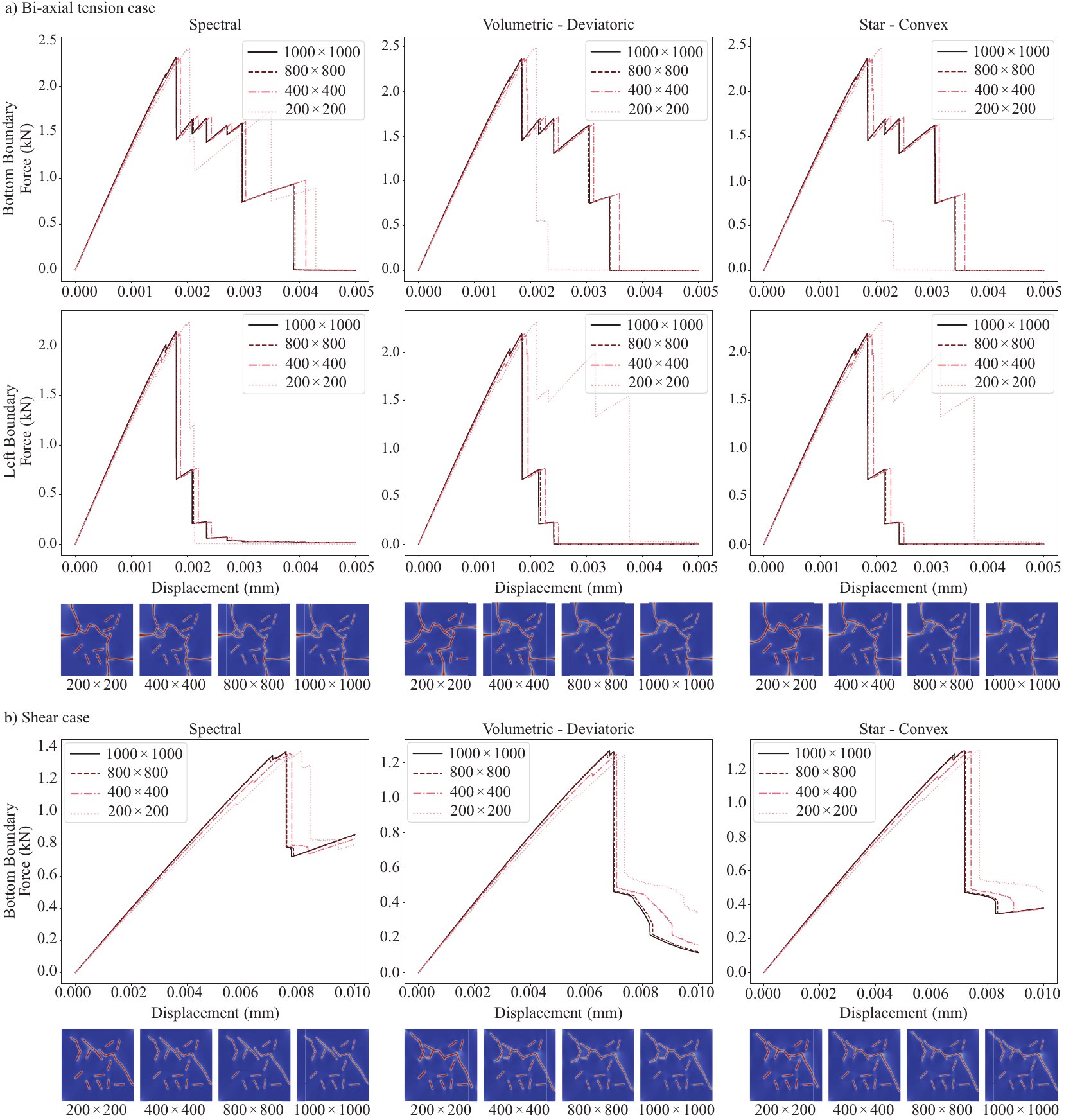}
    \caption{This figure presents the mesh study for a) Bi-axial tension case and b) Shear case with force displacement curves and crack patterns across mesh sizes from a quadrilateral $200\times200$ mesh to $1000\times1000$ mesh. As can be seen the difference between the $800\times800$ mesh and $1000\times1000$ mesh is negligible both in the force displacement curve and final crack pattern.}
    \label{fig:mesh-study}
\end{figure}

\section{Additional results of evaluating FNO model on the test dataset}
In Section~\ref{sec:fno-result}, we presented the performance of the FNO model on the sample with median Dice score, here we present four additional samples from the best performing to the worst performing sample.

\begin{figure}[htbp]
    \centering
    \includegraphics[width=0.85\linewidth]{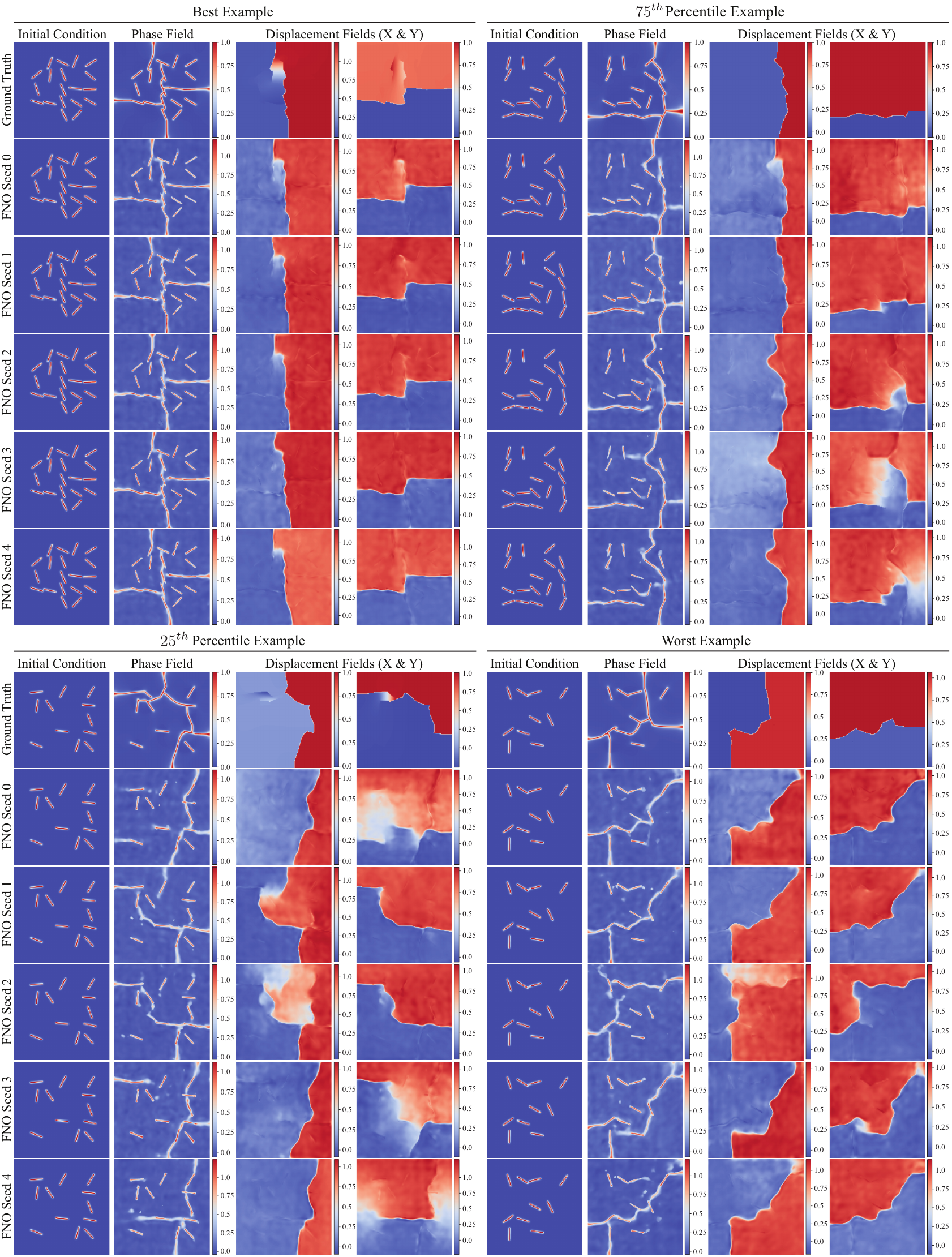}
    \caption{This figure presents the predicted phase field and displacement fields for each FNO model that was trained on the dataset. From the best performing sample with Dice score of 0.8059 to the worst performing sample with Dice score of 0.5588.}
    \label{fig:appendix-fields-1}
\end{figure}

\begin{figure}[htbp]
    \centering
    \includegraphics[width=1\linewidth]{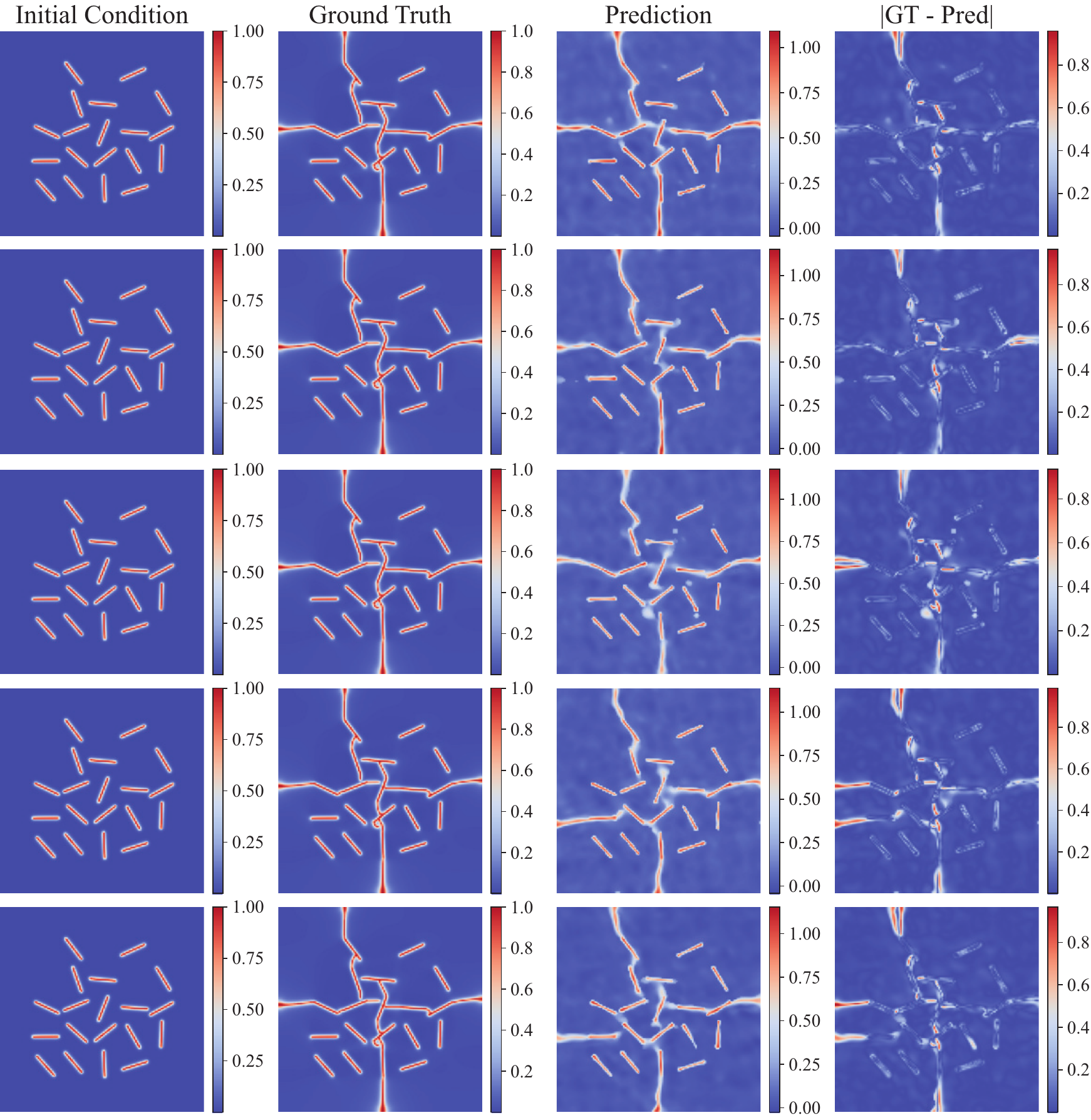}
    \caption{This figure presents the initial condition, ground truth and predicted phase field and the absolute difference of the prediction of the phase field channel from the ground truth for each FNO seed.}
    \label{fig:mse-diff-1}
\end{figure}

\begin{figure}[htbp]
    \centering
    \includegraphics[width=0.8\linewidth]{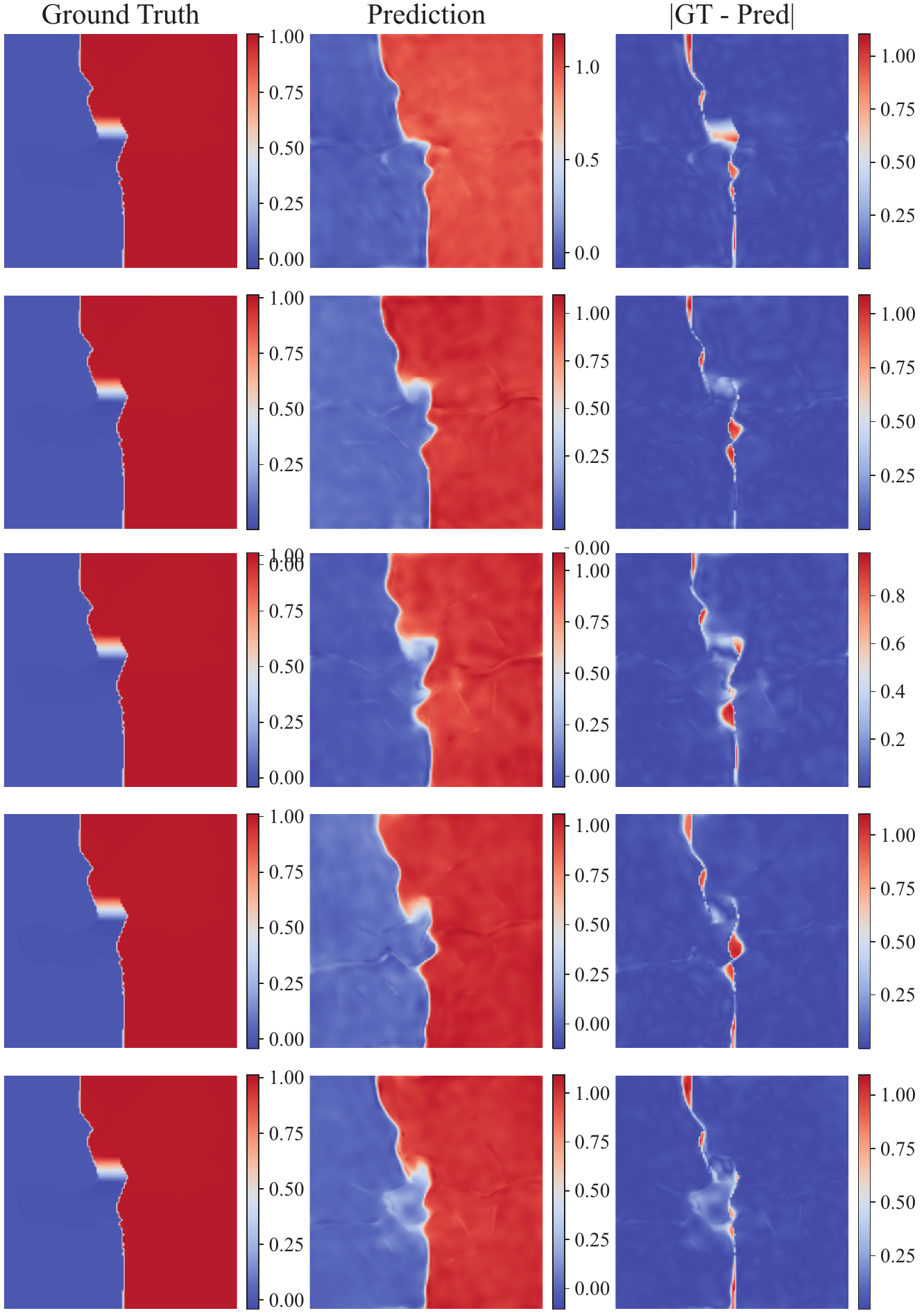}
    \caption{This figure presents the ground truth and the predicted X component of the displacement field and the absolute difference of the prediction of the X component of the displacement field channel from the ground truth for each FNO seed.}
    \label{fig:mse-diff-2}
\end{figure}

\begin{figure}[htbp]
    \centering
    \includegraphics[width=0.8\linewidth]{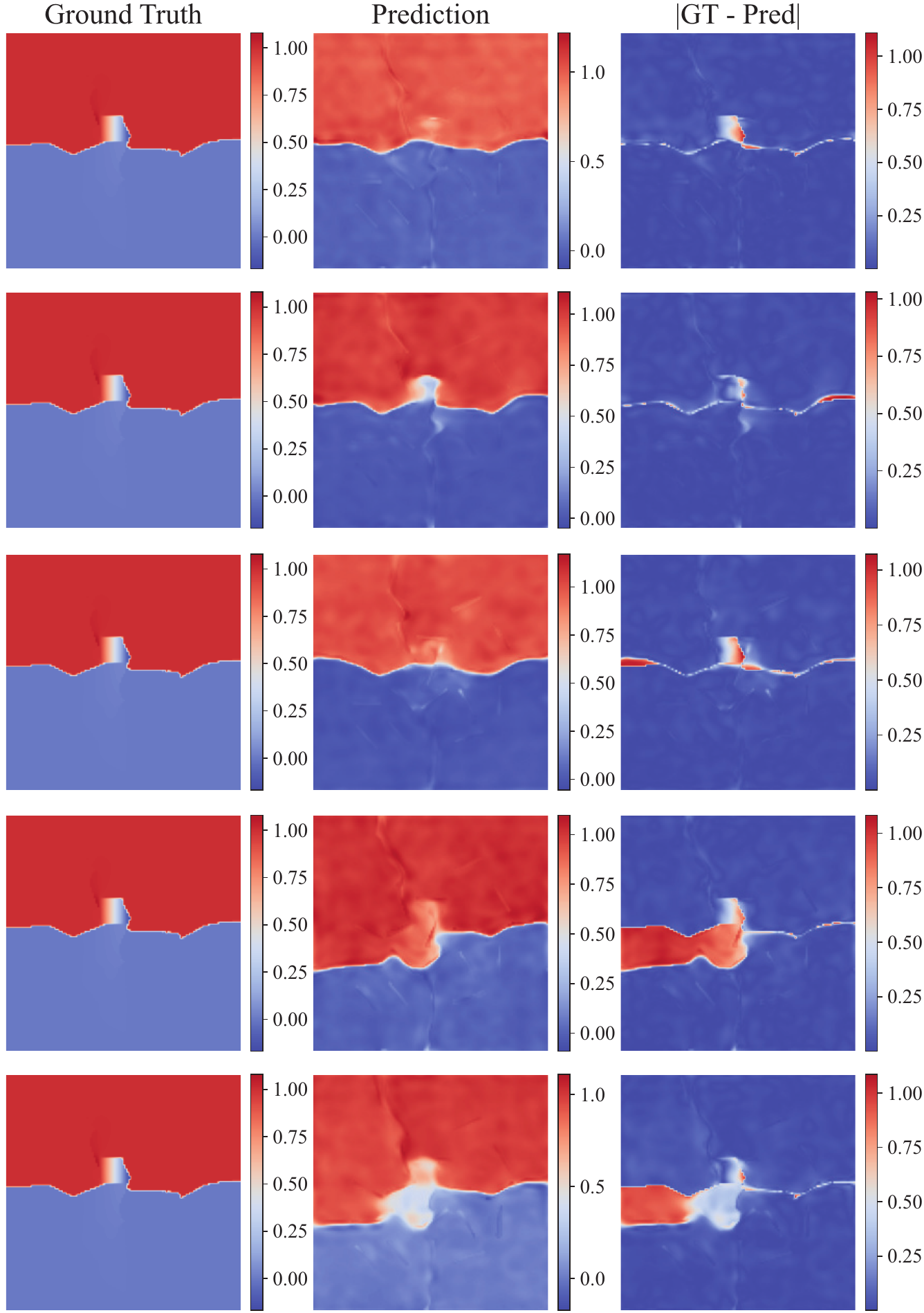}
    \caption{This figure presents the ground truth and the predicted Y component of the displacement field and the absolute difference of the prediction of the Y component of the displacement field channel from the ground truth for each FNO seed.}
    \label{fig:mse-diff-3}
\end{figure}

In Fig.~\ref{fig:mse-diff-1}-\ref{fig:mse-diff-3}, we show the ground truth, the prediction and the absolute difference of the prediction of each channel from the ground truth for each FNO seed.

\newpage
\bibliographystyle{unsrtnat}
\bibliography{main}

\begin{thebibliography}{132}
\providecommand{\natexlab}[1]{#1}
\providecommand{\url}[1]{\texttt{#1}}
\expandafter\ifx\csname urlstyle\endcsname\relax
  \providecommand{\doi}[1]{doi: #1}\else
  \providecommand{\doi}{doi: \begingroup \urlstyle{rm}\Url}\fi

\bibitem[Forrester and Keane(2009)]{forrester2009recent}
Alexander~IJ Forrester and Andy~J Keane.
\newblock Recent advances in surrogate-based optimization.
\newblock \emph{Progress in aerospace sciences}, 45\penalty0 (1-3):\penalty0 50--79, 2009.

\bibitem[Lu and Tartakovsky(2023)]{lu2023drips}
Hannah Lu and Daniel~M Tartakovsky.
\newblock Drips: A framework for dimension reduction and interpolation in parameter space.
\newblock \emph{Journal of Computational Physics}, 493:\penalty0 112455, 2023.

\bibitem[Lu and Tartakovsky(2024)]{lu2024data}
Hannah Lu and Daniel~M Tartakovsky.
\newblock Data-driven models of nonautonomous systems.
\newblock \emph{Journal of Computational Physics}, 507:\penalty0 112976, 2024.

\bibitem[Liu et~al.(2020)Liu, Shang, Ouyang, and Widanage]{liu2020data}
Kailong Liu, Yunlong Shang, Quan Ouyang, and Widanalage~Dhammika Widanage.
\newblock A data-driven approach with uncertainty quantification for predicting future capacities and remaining useful life of lithium-ion battery.
\newblock \emph{IEEE Transactions on Industrial Electronics}, 68\penalty0 (4):\penalty0 3170--3180, 2020.

\bibitem[Boyce et~al.(2022)Boyce, Mart{\'\i}nez-Pa{\~n}eda, Wade, Zhang, Bailey, Heenan, Brett, and Shearing]{boyce2022cracking}
Adam~M Boyce, Emilio Mart{\'\i}nez-Pa{\~n}eda, Aaron Wade, Ye~Shui Zhang, Josh~J Bailey, Thomas~MM Heenan, Dan~JL Brett, and Paul~R Shearing.
\newblock Cracking predictions of lithium-ion battery electrodes by x-ray computed tomography and modelling.
\newblock \emph{Journal of Power Sources}, 526:\penalty0 231119, 2022.

\bibitem[Mo et~al.(2019)Mo, Zhu, Zabaras, Shi, and Wu]{mo2019deep}
Shaoxing Mo, Yinhao Zhu, Nicholas Zabaras, Xiaoqing Shi, and Jichun Wu.
\newblock Deep convolutional encoder-decoder networks for uncertainty quantification of dynamic multiphase flow in heterogeneous media.
\newblock \emph{Water Resources Research}, 55\penalty0 (1):\penalty0 703--728, 2019.

\bibitem[Bhatnagar et~al.(2019)Bhatnagar, Afshar, Pan, Duraisamy, and Kaushik]{bhatnagar2019prediction}
Saakaar Bhatnagar, Yaser Afshar, Shaowu Pan, Karthik Duraisamy, and Shailendra Kaushik.
\newblock Prediction of aerodynamic flow fields using convolutional neural networks.
\newblock \emph{Computational Mechanics}, 64:\penalty0 525--545, 2019.

\bibitem[Prachaseree and Lejeune(2022)]{prachaseree2022learning}
Peerasait Prachaseree and Emma Lejeune.
\newblock Learning mechanically driven emergent behavior with message passing neural networks.
\newblock \emph{Computers \& Structures}, 270:\penalty0 106825, 2022.

\bibitem[Zhao et~al.(2024)Zhao, Li, Zhou, Attar, Pfaff, and Li]{zhao2024review}
Yingxue Zhao, Haoran Li, Haosu Zhou, Hamid~Reza Attar, Tobias Pfaff, and Nan Li.
\newblock A review of graph neural network applications in mechanics-related domains.
\newblock \emph{Artificial Intelligence Review}, 57\penalty0 (11):\penalty0 315, 2024.

\bibitem[Raissi et~al.(2019)Raissi, Perdikaris, and Karniadakis]{raissi2019physics}
Maziar Raissi, Paris Perdikaris, and George~E Karniadakis.
\newblock Physics-informed neural networks: A deep learning framework for solving forward and inverse problems involving nonlinear partial differential equations.
\newblock \emph{Journal of Computational physics}, 378:\penalty0 686--707, 2019.

\bibitem[Mao et~al.(2020)Mao, Jagtap, and Karniadakis]{mao2020physics}
Zhiping Mao, Ameya~D Jagtap, and George~Em Karniadakis.
\newblock Physics-informed neural networks for high-speed flows.
\newblock \emph{Computer Methods in Applied Mechanics and Engineering}, 360:\penalty0 112789, 2020.

\bibitem[Sahli~Costabal et~al.(2020)Sahli~Costabal, Yang, Perdikaris, Hurtado, and Kuhl]{sahli2020physics}
Francisco Sahli~Costabal, Yibo Yang, Paris Perdikaris, Daniel~E Hurtado, and Ellen Kuhl.
\newblock Physics-informed neural networks for cardiac activation mapping.
\newblock \emph{Frontiers in Physics}, 8:\penalty0 42, 2020.

\bibitem[Haghighat et~al.(2021)Haghighat, Raissi, Moure, Gomez, and Juanes]{haghighat2021physics}
Ehsan Haghighat, Maziar Raissi, Adrian Moure, Hector Gomez, and Ruben Juanes.
\newblock A physics-informed deep learning framework for inversion and surrogate modeling in solid mechanics.
\newblock \emph{Computer Methods in Applied Mechanics and Engineering}, 379:\penalty0 113741, 2021.

\bibitem[Fang and Zhan(2019)]{fang2019deep}
Zhiwei Fang and Justin Zhan.
\newblock Deep physical informed neural networks for metamaterial design.
\newblock \emph{Ieee Access}, 8:\penalty0 24506--24513, 2019.

\bibitem[Misyris et~al.(2020)Misyris, Venzke, and Chatzivasileiadis]{misyris2020physics}
George~S Misyris, Andreas Venzke, and Spyros Chatzivasileiadis.
\newblock Physics-informed neural networks for power systems.
\newblock In \emph{2020 IEEE power \& energy society general meeting (PESGM)}, pages 1--5. IEEE, 2020.

\bibitem[Lu et~al.(2019)Lu, Jin, and Karniadakis]{lu2019deeponet}
Lu~Lu, Pengzhan Jin, and George~Em Karniadakis.
\newblock Deeponet: Learning nonlinear operators for identifying differential equations based on the universal approximation theorem of operators.
\newblock \emph{arXiv preprint arXiv:1910.03193}, 2019.

\bibitem[Li et~al.(2020{\natexlab{a}})Li, Kovachki, Azizzadenesheli, Liu, Bhattacharya, Stuart, and Anandkumar]{li2020neural}
Zongyi Li, Nikola Kovachki, Kamyar Azizzadenesheli, Burigede Liu, Kaushik Bhattacharya, Andrew Stuart, and Anima Anandkumar.
\newblock Neural operator: Graph kernel network for partial differential equations.
\newblock \emph{arXiv preprint arXiv:2003.03485}, 2020{\natexlab{a}}.

\bibitem[Patel et~al.(2021)Patel, Trask, Wood, and Cyr]{patel2021physics}
Ravi~G Patel, Nathaniel~A Trask, Mitchell~A Wood, and Eric~C Cyr.
\newblock A physics-informed operator regression framework for extracting data-driven continuum models.
\newblock \emph{Computer Methods in Applied Mechanics and Engineering}, 373:\penalty0 113500, 2021.

\bibitem[Wang et~al.(2021{\natexlab{a}})Wang, Wang, and Perdikaris]{wang2021learning}
Sifan Wang, Hanwen Wang, and Paris Perdikaris.
\newblock Learning the solution operator of parametric partial differential equations with physics-informed deeponets.
\newblock \emph{Science advances}, 7\penalty0 (40):\penalty0 eabi8605, 2021{\natexlab{a}}.

\bibitem[Kiyani et~al.(2025{\natexlab{a}})Kiyani, Manav, Kadivar, De~Lorenzis, and Karniadakis]{kiyani2025predicting}
Elham Kiyani, Manav Manav, Nikhil Kadivar, Laura De~Lorenzis, and George~Em Karniadakis.
\newblock Predicting crack nucleation and propagation in brittle materials using deep operator networks with diverse trunk architectures.
\newblock \emph{Computer Methods in Applied Mechanics and Engineering}, 441:\penalty0 117984, 2025{\natexlab{a}}.

\bibitem[Goswami et~al.(2023)Goswami, Bora, Yu, and Karniadakis]{goswami2023physics}
Somdatta Goswami, Aniruddha Bora, Yue Yu, and George~Em Karniadakis.
\newblock Physics-informed deep neural operator networks.
\newblock In \emph{Machine learning in modeling and simulation: methods and applications}, pages 219--254. Springer, 2023.

\bibitem[Utkarsh et~al.(2025)Utkarsh, Cai, Edelman, Gomez-Bombarelli, and Rackauckas]{utkarsh2025physics}
Utkarsh Utkarsh, Pengfei Cai, Alan Edelman, Rafael Gomez-Bombarelli, and Christopher~Vincent Rackauckas.
\newblock Physics-constrained flow matching: Sampling generative models with hard constraints.
\newblock \emph{arXiv preprint arXiv:2506.04171}, 2025.

\bibitem[Hansen et~al.(2023)Hansen, Maddix, Alizadeh, Gupta, and Mahoney]{hansen2023learning}
Derek Hansen, Danielle~C Maddix, Shima Alizadeh, Gaurav Gupta, and Michael~W Mahoney.
\newblock Learning physical models that can respect conservation laws.
\newblock In \emph{International Conference on Machine Learning}, pages 12469--12510. PMLR, 2023.

\bibitem[Wang et~al.(2021{\natexlab{b}})Wang, Teng, and Perdikaris]{wang2021understanding}
Sifan Wang, Yujun Teng, and Paris Perdikaris.
\newblock Understanding and mitigating gradient flow pathologies in physics-informed neural networks.
\newblock \emph{SIAM Journal on Scientific Computing}, 43\penalty0 (5):\penalty0 A3055--A3081, 2021{\natexlab{b}}.

\bibitem[Berrone et~al.(2023)Berrone, Canuto, Pintore, and Sukumar]{berrone2023enforcing}
Stefano Berrone, Claudio Canuto, Moreno Pintore, and Natarajan Sukumar.
\newblock Enforcing dirichlet boundary conditions in physics-informed neural networks and variational physics-informed neural networks.
\newblock \emph{Heliyon}, 9\penalty0 (8), 2023.

\bibitem[Rahaman et~al.(2019)Rahaman, Baratin, Arpit, Draxler, Lin, Hamprecht, Bengio, and Courville]{rahaman2019spectral}
Nasim Rahaman, Aristide Baratin, Devansh Arpit, Felix Draxler, Min Lin, Fred Hamprecht, Yoshua Bengio, and Aaron Courville.
\newblock On the spectral bias of neural networks.
\newblock In \emph{International conference on machine learning}, pages 5301--5310. PMLR, 2019.

\bibitem[Khodakarami et~al.(2025)Khodakarami, Oommen, Bora, and Karniadakis]{khodakarami2025mitigating}
Siavash Khodakarami, Vivek Oommen, Aniruddha Bora, and George~Em Karniadakis.
\newblock Mitigating spectral bias in neural operators via high-frequency scaling for physical systems.
\newblock \emph{arXiv preprint arXiv:2503.13695}, 2025.

\bibitem[Dauphin et~al.(2014)Dauphin, Pascanu, Gulcehre, Cho, Ganguli, and Bengio]{dauphin2014identifying}
Yann~N Dauphin, Razvan Pascanu, Caglar Gulcehre, Kyunghyun Cho, Surya Ganguli, and Yoshua Bengio.
\newblock Identifying and attacking the saddle point problem in high-dimensional non-convex optimization.
\newblock \emph{Advances in neural information processing systems}, 27, 2014.

\bibitem[Pascanu et~al.(2014)Pascanu, Dauphin, Ganguli, and Bengio]{pascanu2014saddle}
Razvan Pascanu, Yann~N Dauphin, Surya Ganguli, and Yoshua Bengio.
\newblock On the saddle point problem for non-convex optimization.
\newblock \emph{arXiv preprint arXiv:1405.4604}, 2014.

\bibitem[Kiyani et~al.(2025{\natexlab{b}})Kiyani, Shukla, Urb{\'a}n, Darbon, and Karniadakis]{kiyani2025optimizer}
Elham Kiyani, Khemraj Shukla, Jorge~F Urb{\'a}n, J{\'e}r{\^o}me Darbon, and George~Em Karniadakis.
\newblock Which optimizer works best for physics-informed neural networks and kolmogorov-arnold networks?
\newblock \emph{arXiv preprint arXiv:2501.16371}, 2025{\natexlab{b}}.

\bibitem[Frankle and Carbin(2018)]{frankle2018lottery}
Jonathan Frankle and Michael Carbin.
\newblock The lottery ticket hypothesis: Finding sparse, trainable neural networks.
\newblock \emph{arXiv preprint arXiv:1803.03635}, 2018.

\bibitem[Glorot and Bengio(2010)]{glorot2010understanding}
Xavier Glorot and Yoshua Bengio.
\newblock Understanding the difficulty of training deep feedforward neural networks.
\newblock In \emph{Proceedings of the thirteenth international conference on artificial intelligence and statistics}, pages 249--256. JMLR Workshop and Conference Proceedings, 2010.

\bibitem[Bouthillier et~al.(2021)Bouthillier, Delaunay, Bronzi, Trofimov, Nichyporuk, Szeto, Mohammadi~Sepahvand, Raff, Madan, Voleti, et~al.]{bouthillier2021accounting}
Xavier Bouthillier, Pierre Delaunay, Mirko Bronzi, Assya Trofimov, Brennan Nichyporuk, Justin Szeto, Nazanin Mohammadi~Sepahvand, Edward Raff, Kanika Madan, Vikram Voleti, et~al.
\newblock Accounting for variance in machine learning benchmarks.
\newblock \emph{Proceedings of Machine Learning and Systems}, 3:\penalty0 747--769, 2021.

\bibitem[Lejeune(2020)]{lejeune2020mechanical}
Emma Lejeune.
\newblock Mechanical mnist: A benchmark dataset for mechanical metamodels.
\newblock \emph{Extreme Mechanics Letters}, 36:\penalty0 100659, 2020.

\bibitem[Nguyen and Lejeune(2023)]{nguyen2023mechanical}
Quan Nguyen and Emma Lejeune.
\newblock Mechanical mnist--unsupervised learning dataset.
\newblock 2023.

\bibitem[Kobeissi and Lejeune(2022)]{kobeissi2022mechanical}
Hiba Kobeissi and Emma Lejeune.
\newblock Mechanical mnist-cahn-hilliard.
\newblock 2022.

\bibitem[Ohana et~al.(2024)Ohana, McCabe, Meyer, Morel, Agocs, Beneitez, Berger, Burkhart, Dalziel, Fielding, et~al.]{ohana2024well}
Ruben Ohana, Michael McCabe, Lucas Meyer, Rudy Morel, Fruzsina Agocs, Miguel Beneitez, Marsha Berger, Blakesly Burkhart, Stuart Dalziel, Drummond Fielding, et~al.
\newblock The well: a large-scale collection of diverse physics simulations for machine learning.
\newblock \emph{Advances in Neural Information Processing Systems}, 37:\penalty0 44989--45037, 2024.

\bibitem[Takamoto et~al.(2022)Takamoto, Praditia, Leiteritz, MacKinlay, Alesiani, Pfl{\"u}ger, and Niepert]{takamoto2022pdebench}
Makoto Takamoto, Timothy Praditia, Raphael Leiteritz, Daniel MacKinlay, Francesco Alesiani, Dirk Pfl{\"u}ger, and Mathias Niepert.
\newblock Pdebench: An extensive benchmark for scientific machine learning.
\newblock \emph{Advances in Neural Information Processing Systems}, 35:\penalty0 1596--1611, 2022.

\bibitem[Woodford et~al.(2010)Woodford, Chiang, and Carter]{woodford2010electrochemical}
William~H Woodford, Yet-Ming Chiang, and W~Craig Carter.
\newblock “electrochemical shock” of intercalation electrodes: a fracture mechanics analysis.
\newblock \emph{Journal of The Electrochemical Society}, 157\penalty0 (10):\penalty0 A1052, 2010.

\bibitem[Klinsmann et~al.(2016)Klinsmann, Rosato, Kamlah, and McMeeking]{klinsmann2016modeling}
Markus Klinsmann, Daniele Rosato, Marc Kamlah, and Robert~M McMeeking.
\newblock Modeling crack growth during li insertion in storage particles using a fracture phase field approach.
\newblock \emph{Journal of the Mechanics and Physics of Solids}, 92:\penalty0 313--344, 2016.

\bibitem[Sukumar et~al.(2000)Sukumar, Mo{\"e}s, Moran, and Belytschko]{sukumar2000extended}
Natarajan Sukumar, Nicolas Mo{\"e}s, Brian Moran, and Ted Belytschko.
\newblock Extended finite element method for three-dimensional crack modelling.
\newblock \emph{International journal for numerical methods in engineering}, 48\penalty0 (11):\penalty0 1549--1570, 2000.

\bibitem[Talebi et~al.(2013)Talebi, Silani, Bordas, Kerfriden, and Rabczuk]{talebi2013molecular}
Hossein Talebi, Mohammad Silani, St{\'e}phane~PA Bordas, Pierre Kerfriden, and Timon Rabczuk.
\newblock Molecular dynamics/xfem coupling by a three-dimensional extended bridging domain with applications to dynamic brittle fracture.
\newblock \emph{International Journal for Multiscale Computational Engineering}, 11\penalty0 (6), 2013.

\bibitem[Lorentz and Andrieux(1999)]{lorentz1999variational}
Eric Lorentz and St{\'e}phane Andrieux.
\newblock A variational formulation for nonlocal damage models.
\newblock \emph{International journal of plasticity}, 15\penalty0 (2):\penalty0 119--138, 1999.

\bibitem[Goswami et~al.(2020{\natexlab{a}})Goswami, Anitescu, Chakraborty, and Rabczuk]{goswami2020transfer}
Somdatta Goswami, Cosmin Anitescu, Souvik Chakraborty, and Timon Rabczuk.
\newblock Transfer learning enhanced physics informed neural network for phase-field modeling of fracture.
\newblock \emph{Theoretical and Applied Fracture Mechanics}, 106:\penalty0 102447, 2020{\natexlab{a}}.

\bibitem[Manav et~al.(2024)Manav, Molinaro, Mishra, and De~Lorenzis]{manav2024phase}
Manav Manav, Roberto Molinaro, Siddhartha Mishra, and Laura De~Lorenzis.
\newblock Phase-field modeling of fracture with physics-informed deep learning.
\newblock \emph{Computer Methods in Applied Mechanics and Engineering}, 429:\penalty0 117104, 2024.

\bibitem[Gerasimov et~al.(2020)Gerasimov, R{\"o}mer, Vond{\v{r}}ejc, Matthies, and De~Lorenzis]{gerasimov2020stochastic}
Tymofiy Gerasimov, Ulrich R{\"o}mer, Jaroslav Vond{\v{r}}ejc, Hermann~G Matthies, and Laura De~Lorenzis.
\newblock Stochastic phase-field modeling of brittle fracture: computing multiple crack patterns and their probabilities.
\newblock \emph{Computer Methods in Applied Mechanics and Engineering}, 372:\penalty0 113353, 2020.

\bibitem[Zhang et~al.(2024)Zhang, Dolbow, and Guilleminot]{zhang2024representing}
Hao Zhang, John~E Dolbow, and Johann Guilleminot.
\newblock Representing model uncertainties in brittle fracture simulations.
\newblock \emph{Computer Methods in Applied Mechanics and Engineering}, 418:\penalty0 116575, 2024.

\bibitem[Wu et~al.(2020)Wu, Nguyen, Nguyen, Sutula, Sinaie, and Bordas]{wu2020phase}
Jian-Ying Wu, Vinh~Phu Nguyen, Chi~Thanh Nguyen, Danas Sutula, Sina Sinaie, and St{\'e}phane~PA Bordas.
\newblock Phase-field modeling of fracture.
\newblock \emph{Advances in applied mechanics}, 53:\penalty0 1--183, 2020.

\bibitem[Miehe et~al.(2010)Miehe, Hofacker, and Welschinger]{miehe2010phase}
Christian Miehe, Martina Hofacker, and Fabian Welschinger.
\newblock A phase field model for rate-independent crack propagation: Robust algorithmic implementation based on operator splits.
\newblock \emph{Computer Methods in Applied Mechanics and Engineering}, 199\penalty0 (45-48):\penalty0 2765--2778, 2010.

\bibitem[Ambati et~al.(2015)Ambati, Gerasimov, and De~Lorenzis]{ambati2015review}
Marreddy Ambati, Tymofiy Gerasimov, and Laura De~Lorenzis.
\newblock A review on phase-field models of brittle fracture and a new fast hybrid formulation.
\newblock \emph{Computational Mechanics}, 55:\penalty0 383--405, 2015.

\bibitem[Goswami et~al.(2020{\natexlab{b}})Goswami, Anitescu, and Rabczuk]{goswami2020adaptive}
Somdatta Goswami, Cosmin Anitescu, and Timon Rabczuk.
\newblock Adaptive fourth-order phase field analysis using deep energy minimization.
\newblock \emph{Theoretical and Applied Fracture Mechanics}, 107:\penalty0 102527, 2020{\natexlab{b}}.

\bibitem[Goswami et~al.(2022)Goswami, Yin, Yu, and Karniadakis]{goswami2022physics}
Somdatta Goswami, Minglang Yin, Yue Yu, and George~Em Karniadakis.
\newblock A physics-informed variational deeponet for predicting crack path in quasi-brittle materials.
\newblock \emph{Computer Methods in Applied Mechanics and Engineering}, 391:\penalty0 114587, 2022.

\bibitem[Ghaffari~Motlagh et~al.(2023)Ghaffari~Motlagh, Jimack, and de~Borst]{ghaffari2023deep}
Yousef Ghaffari~Motlagh, Peter~K Jimack, and Ren{\'e} de~Borst.
\newblock Deep learning phase-field model for brittle fractures.
\newblock \emph{International Journal for Numerical Methods in Engineering}, 124\penalty0 (3):\penalty0 620--638, 2023.

\bibitem[Tripathy and Bilionis(2018)]{tripathy2018deep}
Rohit~K Tripathy and Ilias Bilionis.
\newblock Deep uq: Learning deep neural network surrogate models for high dimensional uncertainty quantification.
\newblock \emph{Journal of computational physics}, 375:\penalty0 565--588, 2018.

\bibitem[Graves(2011)]{graves2011practical}
Alex Graves.
\newblock Practical variational inference for neural networks.
\newblock \emph{Advances in neural information processing systems}, 24, 2011.

\bibitem[Yang et~al.(2022)Yang, Kissas, and Perdikaris]{yang2022scalable}
Yibo Yang, Georgios Kissas, and Paris Perdikaris.
\newblock Scalable uncertainty quantification for deep operator networks using randomized priors.
\newblock \emph{Computer Methods in Applied Mechanics and Engineering}, 399:\penalty0 115399, 2022.

\bibitem[Fort et~al.(2019)Fort, Hu, and Lakshminarayanan]{fort2019deep}
Stanislav Fort, Huiyi Hu, and Balaji Lakshminarayanan.
\newblock Deep ensembles: A loss landscape perspective.
\newblock \emph{arXiv preprint arXiv:1912.02757}, 2019.

\bibitem[Psaros et~al.(2023)Psaros, Meng, Zou, Guo, and Karniadakis]{psaros2023uncertainty}
Apostolos~F Psaros, Xuhui Meng, Zongren Zou, Ling Guo, and George~Em Karniadakis.
\newblock Uncertainty quantification in scientific machine learning: Methods, metrics, and comparisons.
\newblock \emph{Journal of Computational Physics}, 477:\penalty0 111902, 2023.

\bibitem[Mohammadzadeh and Lejeune(2021)]{mohammadzadeh2021mechanical}
Saeed Mohammadzadeh and Emma Lejeune.
\newblock Mechanical mnist crack path.
\newblock 2021.

\bibitem[Anderson and Anderson(2005)]{anderson2005fracture}
Ted~L Anderson and Ted~L Anderson.
\newblock \emph{Fracture mechanics: fundamentals and applications}.
\newblock CRC press, 2005.

\bibitem[Griffith(1921)]{griffith1921vi}
Alan~Arnold Griffith.
\newblock Vi. the phenomena of rupture and flow in solids.
\newblock \emph{Philosophical transactions of the royal society of london. Series A, containing papers of a mathematical or physical character}, 221\penalty0 (582-593):\penalty0 163--198, 1921.

\bibitem[Irwin(1957)]{irwin1957analysis}
George~R Irwin.
\newblock Analysis of stresses and strains near the end of a crack traversing a plate.
\newblock 1957.

\bibitem[Dugdale(1960)]{dugdale1960yielding}
Donald~S Dugdale.
\newblock Yielding of steel sheets containing slits.
\newblock \emph{Journal of the Mechanics and Physics of Solids}, 8\penalty0 (2):\penalty0 100--104, 1960.

\bibitem[Kuhn and M{\"u}ller(2010)]{kuhn2010continuum}
Charlotte Kuhn and Ralf M{\"u}ller.
\newblock A continuum phase field model for fracture.
\newblock \emph{Engineering fracture mechanics}, 77\penalty0 (18):\penalty0 3625--3634, 2010.

\bibitem[Amor et~al.(2009)Amor, Marigo, and Maurini]{amor2009regularized}
Hanen Amor, Jean-Jacques Marigo, and Corrado Maurini.
\newblock Regularized formulation of the variational brittle fracture with unilateral contact: Numerical experiments.
\newblock \emph{Journal of the Mechanics and Physics of Solids}, 57\penalty0 (8):\penalty0 1209--1229, 2009.

\bibitem[Bourdin et~al.(2008)Bourdin, Francfort, and Marigo]{bourdin2008variational}
Blaise Bourdin, Gilles~A Francfort, and Jean-Jacques Marigo.
\newblock The variational approach to fracture.
\newblock \emph{Journal of elasticity}, 91:\penalty0 5--148, 2008.

\bibitem[Wu(2017)]{wu2017unified}
Jian-Ying Wu.
\newblock A unified phase-field theory for the mechanics of damage and quasi-brittle failure.
\newblock \emph{Journal of the Mechanics and Physics of Solids}, 103:\penalty0 72--99, 2017.

\bibitem[Vicentini et~al.(2024)Vicentini, Zolesi, Carrara, Maurini, and De~Lorenzis]{vicentini2024energy}
Francesco Vicentini, Camilla Zolesi, Pietro Carrara, Corrado Maurini, and Laura De~Lorenzis.
\newblock On the energy decomposition in variational phase-field models for brittle fracture under multi-axial stress states.
\newblock \emph{International Journal of Fracture}, pages 1--27, 2024.

\bibitem[Kumar et~al.(2018)Kumar, Francfort, and Lopez-Pamies]{kumar2018fracture}
Aditya Kumar, Gilles~A Francfort, and Oscar Lopez-Pamies.
\newblock Fracture and healing of elastomers: A phase-transition theory and numerical implementation.
\newblock \emph{Journal of the Mechanics and Physics of Solids}, 112:\penalty0 523--551, 2018.

\bibitem[Kamarei et~al.(2024)Kamarei, Kumar, and Lopez-Pamies]{kamarei2024poker}
Farhad Kamarei, Aditya Kumar, and Oscar Lopez-Pamies.
\newblock The poker-chip experiments of synthetic elastomers explained.
\newblock \emph{Journal of the Mechanics and Physics of Solids}, 188:\penalty0 105683, 2024.

\bibitem[Liu and Kumar(2025)]{liu2025emergence}
Chang Liu and Aditya Kumar.
\newblock Emergence of tension--compression asymmetry from a complete phase-field approach to brittle fracture.
\newblock \emph{International Journal of Solids and Structures}, 309:\penalty0 113170, 2025.

\bibitem[Senthilnathan(2025)]{senthilnathan2025construction}
Chockalingam Senthilnathan.
\newblock On the construction of explicit analytical driving forces for crack nucleation in the phase field approach to brittle fracture with application to mohr-coulomb and drucker-prager strength surfaces.
\newblock \emph{Journal of Applied Mechanics}, pages 1--12, 2025.

\bibitem[Freddi and Royer-Carfagni(2010)]{freddi2010regularized}
Francesco Freddi and Gianni Royer-Carfagni.
\newblock Regularized variational theories of fracture: a unified approach.
\newblock \emph{Journal of the Mechanics and Physics of Solids}, 58\penalty0 (8):\penalty0 1154--1174, 2010.

\bibitem[De~Lorenzis and Maurini(2022)]{de2022nucleation}
Laura De~Lorenzis and Corrado Maurini.
\newblock Nucleation under multi-axial loading in variational phase-field models of brittle fracture.
\newblock \emph{International Journal of Fracture}, 237\penalty0 (1):\penalty0 61--81, 2022.

\bibitem[Vajari et~al.(2023)Vajari, Neuner, Arunachala, and Linder]{vajari2023investigation}
Sina~Abrari Vajari, Matthias Neuner, Prajwal~Kammardi Arunachala, and Christian Linder.
\newblock Investigation of driving forces in a phase field approach to mixed mode fracture of concrete.
\newblock \emph{Computer Methods in Applied Mechanics and Engineering}, 417:\penalty0 116404, 2023.

\bibitem[Tang et~al.(2019)Tang, Zhang, Guo, Guo, and Liu]{tang2019phase}
Shan Tang, Gang Zhang, Tian~Fu Guo, Xu~Guo, and Wing~Kam Liu.
\newblock Phase field modeling of fracture in nonlinearly elastic solids via energy decomposition.
\newblock \emph{Computer Methods in Applied Mechanics and Engineering}, 347:\penalty0 477--494, 2019.

\bibitem[van Dijk et~al.(2020)van Dijk, Espadas-Escalante, and Isaksson]{van2020strain}
Nico~P van Dijk, Juan~Jos{\'e} Espadas-Escalante, and Per Isaksson.
\newblock Strain energy density decompositions in phase-field fracture theories for orthotropy and anisotropy.
\newblock \emph{International Journal of Solids and Structures}, 196:\penalty0 140--153, 2020.

\bibitem[Wu and Nguyen(2018)]{wu2018length}
Jian-Ying Wu and Vinh~Phu Nguyen.
\newblock A length scale insensitive phase-field damage model for brittle fracture.
\newblock \emph{Journal of the Mechanics and Physics of Solids}, 119:\penalty0 20--42, 2018.

\bibitem[Kumar et~al.(2022)Kumar, Ravi-Chandar, and Lopez-Pamies]{kumar2022revisited}
Aditya Kumar, K~Ravi-Chandar, and Oscar Lopez-Pamies.
\newblock The revisited phase-field approach to brittle fracture: application to indentation and notch problems.
\newblock \emph{International Journal of Fracture}, 237\penalty0 (1):\penalty0 83--100, 2022.

\bibitem[Borden et~al.(2012)Borden, Verhoosel, Scott, Hughes, and Landis]{borden2012phase}
Michael~J Borden, Clemens~V Verhoosel, Michael~A Scott, Thomas~JR Hughes, and Chad~M Landis.
\newblock A phase-field description of dynamic brittle fracture.
\newblock \emph{Computer Methods in Applied Mechanics and Engineering}, 217:\penalty0 77--95, 2012.

\bibitem[Mikeli{\'c} et~al.(2015)Mikeli{\'c}, Wheeler, and Wick]{mikelic2015phase}
A~Mikeli{\'c}, Mary~F Wheeler, and Thomas Wick.
\newblock Phase-field modeling of a fluid-driven fracture in a poroelastic medium.
\newblock \emph{Computational Geosciences}, 19:\penalty0 1171--1195, 2015.

\bibitem[Zhou et~al.(2018{\natexlab{a}})Zhou, Zhuang, and Rabczuk]{zhou2018phasePoro}
Shuwei Zhou, Xiaoying Zhuang, and Timon Rabczuk.
\newblock A phase-field modeling approach of fracture propagation in poroelastic media.
\newblock \emph{Engineering Geology}, 240:\penalty0 189--203, 2018{\natexlab{a}}.

\bibitem[Zhou et~al.(2018{\natexlab{b}})Zhou, Zhuang, Zhu, and Rabczuk]{zhou2018phase}
Shuwei Zhou, Xiaoying Zhuang, Hehua Zhu, and Timon Rabczuk.
\newblock Phase field modelling of crack propagation, branching and coalescence in rocks.
\newblock \emph{Theoretical and Applied Fracture Mechanics}, 96:\penalty0 174--192, 2018{\natexlab{b}}.

\bibitem[Zhuang et~al.(2022)Zhuang, Zhou, Huynh, Areias, and Rabczuk]{zhuang2022phase}
X~Zhuang, S~Zhou, GD~Huynh, P~Areias, and T~Rabczuk.
\newblock Phase field modeling and computer implementation: A review.
\newblock \emph{Engineering Fracture Mechanics}, 262:\penalty0 108234, 2022.

\bibitem[Baratta et~al.(2023)Baratta, Dean, Dokken, Habera, HALE, Richardson, Rognes, Scroggs, Sime, and Wells]{baratta2023dolfinx}
Igor~A Baratta, Joseph~P Dean, J{\o}rgen~S Dokken, Michal Habera, Jack HALE, Chris~N Richardson, Marie~E Rognes, Matthew~W Scroggs, Nathan Sime, and Garth~N Wells.
\newblock Dolfinx: the next generation fenics problem solving environment.
\newblock 2023.

\bibitem[Scroggs et~al.(2022{\natexlab{a}})Scroggs, Dokken, Richardson, and Wells]{scroggs2022construction}
Matthew~W Scroggs, J{\o}rgen~S Dokken, Chris~N Richardson, and Garth~N Wells.
\newblock Construction of arbitrary order finite element degree-of-freedom maps on polygonal and polyhedral cell meshes.
\newblock \emph{ACM Transactions on Mathematical Software (TOMS)}, 48\penalty0 (2):\penalty0 1--23, 2022{\natexlab{a}}.

\bibitem[Scroggs et~al.(2022{\natexlab{b}})Scroggs, Baratta, Richardson, and Wells]{scroggs2022basix}
Matthew~W Scroggs, Igor~A Baratta, Chris~N Richardson, and Garth~N Wells.
\newblock Basix: a runtime finite element basis evaluation library.
\newblock \emph{Journal of Open Source Software}, 7\penalty0 (73):\penalty0 3982, 2022{\natexlab{b}}.

\bibitem[Aln{\ae}s et~al.(2014)Aln{\ae}s, Logg, {\O}lgaard, Rognes, and Wells]{alnaes2014unified}
Martin~S Aln{\ae}s, Anders Logg, Kristian~B {\O}lgaard, Marie~E Rognes, and Garth~N Wells.
\newblock Unified form language: A domain-specific language for weak formulations of partial differential equations.
\newblock \emph{ACM Transactions on Mathematical Software (TOMS)}, 40\penalty0 (2):\penalty0 1--37, 2014.

\bibitem[Yu et~al.(2018)]{yu2018deep}
Bing Yu et~al.
\newblock The deep ritz method: a deep learning-based numerical algorithm for solving variational problems.
\newblock \emph{Communications in Mathematics and Statistics}, 6\penalty0 (1):\penalty0 1--12, 2018.

\bibitem[Paszke(2019)]{paszke2019pytorch}
A~Paszke.
\newblock Pytorch: An imperative style, high-performance deep learning library.
\newblock \emph{arXiv preprint arXiv:1912.01703}, 2019.

\bibitem[Biewald(2020)]{wandb}
Lukas Biewald.
\newblock Experiment tracking with weights and biases, 2020.
\newblock URL \url{https://www.wandb.com/}.
\newblock Software available from wandb.com.

\bibitem[Mohammadzadeh et~al.(2023)Mohammadzadeh, Prachaseree, and Lejeune]{mohammadzadeh2023investigating}
Saeed Mohammadzadeh, Peerasait Prachaseree, and Emma Lejeune.
\newblock Investigating deep learning model calibration for classification problems in mechanics.
\newblock \emph{Mechanics of Materials}, 184:\penalty0 104749, 2023.

\bibitem[Zhang et~al.(2022)Zhang, Dao, Karniadakis, and Suresh]{zhang2022analyses}
Enrui Zhang, Ming Dao, George~Em Karniadakis, and Subra Suresh.
\newblock Analyses of internal structures and defects in materials using physics-informed neural networks.
\newblock \emph{Science advances}, 8\penalty0 (7):\penalty0 eabk0644, 2022.

\bibitem[Rasht-Behesht et~al.(2022)Rasht-Behesht, Huber, Shukla, and Karniadakis]{rasht2022physics}
Majid Rasht-Behesht, Christian Huber, Khemraj Shukla, and George~Em Karniadakis.
\newblock Physics-informed neural networks (pinns) for wave propagation and full waveform inversions.
\newblock \emph{Journal of Geophysical Research: Solid Earth}, 127\penalty0 (5):\penalty0 e2021JB023120, 2022.

\bibitem[Kharazmi et~al.(2021)Kharazmi, Zhang, and Karniadakis]{kharazmi2021hp}
Ehsan Kharazmi, Zhongqiang Zhang, and George~Em Karniadakis.
\newblock hp-vpinns: Variational physics-informed neural networks with domain decomposition.
\newblock \emph{Computer Methods in Applied Mechanics and Engineering}, 374:\penalty0 113547, 2021.

\bibitem[Cai et~al.(2021)Cai, Mao, Wang, Yin, and Karniadakis]{cai2021physics}
Shengze Cai, Zhiping Mao, Zhicheng Wang, Minglang Yin, and George~Em Karniadakis.
\newblock Physics-informed neural networks (pinns) for fluid mechanics: A review.
\newblock \emph{Acta Mechanica Sinica}, 37\penalty0 (12):\penalty0 1727--1738, 2021.

\bibitem[Karniadakis et~al.(2021)Karniadakis, Kevrekidis, Lu, Perdikaris, Wang, and Yang]{karniadakis2021physics}
George~Em Karniadakis, Ioannis~G Kevrekidis, Lu~Lu, Paris Perdikaris, Sifan Wang, and Liu Yang.
\newblock Physics-informed machine learning.
\newblock \emph{Nature Reviews Physics}, 3\penalty0 (6):\penalty0 422--440, 2021.

\bibitem[Khodayi-Mehr and Zavlanos(2020)]{khodayi2020varnet}
Reza Khodayi-Mehr and Michael Zavlanos.
\newblock Varnet: Variational neural networks for the solution of partial differential equations.
\newblock In \emph{Learning for dynamics and control}, pages 298--307. PMLR, 2020.

\bibitem[Chakraborty et~al.(2022)Chakraborty, Anitescu, Goswami, Zhuang, and Rabczuk]{chakraborty2022variational}
Ayan Chakraborty, Cosmin Anitescu, Somdatta Goswami, Xiaoying Zhuang, and Timon Rabczuk.
\newblock Variational energy based xpinns for phase field analysis in brittle fracture.
\newblock \emph{arXiv preprint arXiv:2207.02307}, 2022.

\bibitem[Zheng et~al.(2022)Zheng, Li, Qi, Gao, Liu, and Yuan]{zheng2022physics}
Bin Zheng, Tongchun Li, Huijun Qi, Lingang Gao, Xiaoqing Liu, and Li~Yuan.
\newblock Physics-informed machine learning model for computational fracture of quasi-brittle materials without labelled data.
\newblock \emph{International Journal of Mechanical Sciences}, 223:\penalty0 107282, 2022.

\bibitem[Li et~al.(2020{\natexlab{b}})Li, Kovachki, Azizzadenesheli, Liu, Bhattacharya, Stuart, and Anandkumar]{li2020fourier}
Zongyi Li, Nikola Kovachki, Kamyar Azizzadenesheli, Burigede Liu, Kaushik Bhattacharya, Andrew Stuart, and Anima Anandkumar.
\newblock Fourier neural operator for parametric partial differential equations.
\newblock \emph{arXiv preprint arXiv:2010.08895}, 2020{\natexlab{b}}.

\bibitem[Lu et~al.(2021{\natexlab{a}})Lu, Meng, Mao, and Karniadakis]{lu2021deepxde}
Lu~Lu, Xuhui Meng, Zhiping Mao, and George~Em Karniadakis.
\newblock {DeepXDE}: A deep learning library for solving differential equations.
\newblock \emph{SIAM Review}, 63\penalty0 (1):\penalty0 208--228, 2021{\natexlab{a}}.
\newblock \doi{10.1137/19M1274067}.

\bibitem[Liu and Nocedal(1989)]{liu1989limited}
Dong~C Liu and Jorge Nocedal.
\newblock On the limited memory bfgs method for large scale optimization.
\newblock \emph{Mathematical programming}, 45\penalty0 (1):\penalty0 503--528, 1989.

\bibitem[Riedmiller and Braun(1993)]{riedmiller1993direct}
Martin Riedmiller and Heinrich Braun.
\newblock A direct adaptive method for faster backpropagation learning: The rprop algorithm.
\newblock In \emph{IEEE international conference on neural networks}, pages 586--591. IEEE, 1993.

\bibitem[Kovachki et~al.(2023)Kovachki, Li, Liu, Azizzadenesheli, Bhattacharya, Stuart, and Anandkumar]{kovachki2023neural}
Nikola Kovachki, Zongyi Li, Burigede Liu, Kamyar Azizzadenesheli, Kaushik Bhattacharya, Andrew Stuart, and Anima Anandkumar.
\newblock Neural operator: Learning maps between function spaces with applications to pdes.
\newblock \emph{Journal of Machine Learning Research}, 24\penalty0 (89):\penalty0 1--97, 2023.

\bibitem[Lu et~al.(2021{\natexlab{b}})Lu, Jin, Pang, Zhang, and Karniadakis]{lu2021learning}
Lu~Lu, Pengzhan Jin, Guofei Pang, Zhongqiang Zhang, and George~Em Karniadakis.
\newblock Learning nonlinear operators via deeponet based on the universal approximation theorem of operators.
\newblock \emph{Nature machine intelligence}, 3\penalty0 (3):\penalty0 218--229, 2021{\natexlab{b}}.

\bibitem[Shih et~al.(2025)Shih, Peyvan, Zhang, and Karniadakis]{shih2025transformers}
Benjamin Shih, Ahmad Peyvan, Zhongqiang Zhang, and George~Em Karniadakis.
\newblock Transformers as neural operators for solutions of differential equations with finite regularity.
\newblock \emph{Computer Methods in Applied Mechanics and Engineering}, 434:\penalty0 117560, 2025.

\bibitem[Kingma and Ba(2014)]{kingma2014adam}
Diederik~P Kingma and Jimmy Ba.
\newblock Adam: A method for stochastic optimization.
\newblock \emph{arXiv preprint arXiv:1412.6980}, 2014.

\bibitem[Ronneberger et~al.(2015)Ronneberger, Fischer, and Brox]{ronneberger2015u}
Olaf Ronneberger, Philipp Fischer, and Thomas Brox.
\newblock U-net: Convolutional networks for biomedical image segmentation.
\newblock In \emph{Medical image computing and computer-assisted intervention--MICCAI 2015: 18th international conference, Munich, Germany, October 5-9, 2015, proceedings, part III 18}, pages 234--241. Springer, 2015.

\bibitem[Lin et~al.(2017)Lin, Goyal, Girshick, He, and Doll{\'a}r]{lin2017focal}
Tsung-Yi Lin, Priya Goyal, Ross Girshick, Kaiming He, and Piotr Doll{\'a}r.
\newblock Focal loss for dense object detection.
\newblock In \emph{Proceedings of the IEEE international conference on computer vision}, pages 2980--2988, 2017.

\bibitem[Lakshminarayanan et~al.(2017)Lakshminarayanan, Pritzel, and Blundell]{lakshminarayanan2017simple}
Balaji Lakshminarayanan, Alexander Pritzel, and Charles Blundell.
\newblock Simple and scalable predictive uncertainty estimation using deep ensembles.
\newblock \emph{Advances in neural information processing systems}, 30, 2017.

\bibitem[Lam and Suen(1995)]{lam1995optimal}
Louisa Lam and Ching~Y Suen.
\newblock Optimal combinations of pattern classifiers.
\newblock \emph{Pattern Recognition Letters}, 16\penalty0 (9):\penalty0 945--954, 1995.

\bibitem[Ganaie et~al.(2022)Ganaie, Hu, Malik, Tanveer, and Suganthan]{ganaie2022ensemble}
Mudasir~A Ganaie, Minghui Hu, Ashwani~Kumar Malik, Muhammad Tanveer, and Ponnuthurai~N Suganthan.
\newblock Ensemble deep learning: A review.
\newblock \emph{Engineering Applications of Artificial Intelligence}, 115:\penalty0 105151, 2022.

\bibitem[Wolpert(1992)]{wolpert1992stacked}
David~H Wolpert.
\newblock Stacked generalization.
\newblock \emph{Neural networks}, 5\penalty0 (2):\penalty0 241--259, 1992.

\bibitem[Qin et~al.(2024)Qin, Lyu, Peng, Geng, Wang, Tang, Leroyer, Gao, Liu, and Wang]{qin2024toward}
Shaoxiang Qin, Fuyuan Lyu, Wenhui Peng, Dingyang Geng, Ju~Wang, Xing Tang, Sylvie Leroyer, Naiping Gao, Xue Liu, and Liangzhu~Leon Wang.
\newblock Toward a better understanding of fourier neural operators from a spectral perspective.
\newblock \emph{arXiv preprint arXiv:2404.07200}, 2024.

\bibitem[Sharma and Shankar(2024)]{sharma2024ensemble}
Ramansh Sharma and Varun Shankar.
\newblock Ensemble and mixture-of-experts deeponets for operator learning.
\newblock \emph{arXiv preprint arXiv:2405.11907}, 2024.

\bibitem[Huang et~al.(2017)Huang, Li, Pleiss, Liu, Hopcroft, and Weinberger]{huang2017snapshot}
Gao Huang, Yixuan Li, Geoff Pleiss, Zhuang Liu, John~E Hopcroft, and Kilian~Q Weinberger.
\newblock Snapshot ensembles: Train 1, get m for free.
\newblock \emph{arXiv preprint arXiv:1704.00109}, 2017.

\bibitem[Xu et~al.(2019)Xu, Zhang, Luo, Xiao, and Ma]{xu2019frequency}
Zhi-Qin~John Xu, Yaoyu Zhang, Tao Luo, Yanyang Xiao, and Zheng Ma.
\newblock Frequency principle: Fourier analysis sheds light on deep neural networks.
\newblock \emph{arXiv preprint arXiv:1901.06523}, 2019.

\bibitem[Xu et~al.(2025)Xu, Zhang, and Cai]{xu2025understanding}
Zhi-Qin~John Xu, Lulu Zhang, and Wei Cai.
\newblock On understanding and overcoming spectral biases of deep neural network learning methods for solving pdes.
\newblock \emph{arXiv preprint arXiv:2501.09987}, 2025.

\bibitem[Shazeer et~al.(2017)Shazeer, Mirhoseini, Maziarz, Davis, Le, Hinton, and Dean]{shazeer2017outrageously}
Noam Shazeer, Azalia Mirhoseini, Krzysztof Maziarz, Andy Davis, Quoc Le, Geoffrey Hinton, and Jeff Dean.
\newblock Outrageously large neural networks: The sparsely-gated mixture-of-experts layer.
\newblock \emph{arXiv preprint arXiv:1701.06538}, 2017.

\bibitem[Liu et~al.(2022)Liu, Mao, Wu, Feichtenhofer, Darrell, and Xie]{liu2022convnet}
Zhuang Liu, Hanzi Mao, Chao-Yuan Wu, Christoph Feichtenhofer, Trevor Darrell, and Saining Xie.
\newblock A convnet for the 2020s.
\newblock In \emph{Proceedings of the IEEE/CVF conference on computer vision and pattern recognition}, pages 11976--11986, 2022.

\bibitem[Mohammadzadeh and Lejeune(2022)]{mohammadzadeh2022predicting}
Saeed Mohammadzadeh and Emma Lejeune.
\newblock Predicting mechanically driven full-field quantities of interest with deep learning-based metamodels.
\newblock \emph{Extreme Mechanics Letters}, 50:\penalty0 101566, 2022.

\bibitem[Brandstetter et~al.(2022)Brandstetter, Worrall, and Welling]{brandstetter2022message}
Johannes Brandstetter, Daniel Worrall, and Max Welling.
\newblock Message passing neural pde solvers.
\newblock \emph{arXiv preprint arXiv:2202.03376}, 2022.

\bibitem[Ledig et~al.(2017)Ledig, Theis, Husz{\'a}r, Caballero, Cunningham, Acosta, Aitken, Tejani, Totz, Wang, et~al.]{ledig2017photo}
Christian Ledig, Lucas Theis, Ferenc Husz{\'a}r, Jose Caballero, Andrew Cunningham, Alejandro Acosta, Andrew Aitken, Alykhan Tejani, Johannes Totz, Zehan Wang, et~al.
\newblock Photo-realistic single image super-resolution using a generative adversarial network.
\newblock In \emph{Proceedings of the IEEE conference on computer vision and pattern recognition}, pages 4681--4690, 2017.

\bibitem[Cai and Xu(2019)]{cai2019multi}
Wei Cai and Zhi-Qin~John Xu.
\newblock Multi-scale deep neural networks for solving high dimensional pdes.
\newblock \emph{arXiv preprint arXiv:1910.11710}, 2019.

\bibitem[Tancik et~al.(2020)Tancik, Srinivasan, Mildenhall, Fridovich-Keil, Raghavan, Singhal, Ramamoorthi, Barron, and Ng]{tancik2020fourier}
Matthew Tancik, Pratul Srinivasan, Ben Mildenhall, Sara Fridovich-Keil, Nithin Raghavan, Utkarsh Singhal, Ravi Ramamoorthi, Jonathan Barron, and Ren Ng.
\newblock Fourier features let networks learn high frequency functions in low dimensional domains.
\newblock \emph{Advances in neural information processing systems}, 33:\penalty0 7537--7547, 2020.

\bibitem[Bodnar et~al.(2025)Bodnar, Bruinsma, Lucic, Stanley, Allen, Brandstetter, Garvan, Riechert, Weyn, Dong, et~al.]{bodnar2025foundation}
Cristian Bodnar, Wessel~P Bruinsma, Ana Lucic, Megan Stanley, Anna Allen, Johannes Brandstetter, Patrick Garvan, Maik Riechert, Jonathan~A Weyn, Haiyu Dong, et~al.
\newblock A foundation model for the earth system.
\newblock \emph{Nature}, pages 1--8, 2025.

\bibitem[McCabe et~al.(2023)McCabe, Blancard, Parker, Ohana, Cranmer, Bietti, Eickenberg, Golkar, Krawezik, Lanusse, et~al.]{mccabe2023multiple}
Michael McCabe, Bruno R{\'e}galdo-Saint Blancard, Liam~Holden Parker, Ruben Ohana, Miles Cranmer, Alberto Bietti, Michael Eickenberg, Siavash Golkar, Geraud Krawezik, Francois Lanusse, et~al.
\newblock Multiple physics pretraining for physical surrogate models.
\newblock \emph{arXiv preprint arXiv:2310.02994}, 2023.

\bibitem[Wiesner et~al.(2025)Wiesner, Wessling, and Baek]{wiesner2025towards}
Florian Wiesner, Matthias Wessling, and Stephen Baek.
\newblock Towards a physics foundation model.
\newblock \emph{arXiv preprint arXiv:2509.13805}, 2025.

\bibitem[Wang et~al.(2022)Wang, Yu, and Perdikaris]{wang2022and}
Sifan Wang, Xinling Yu, and Paris Perdikaris.
\newblock When and why pinns fail to train: A neural tangent kernel perspective.
\newblock \emph{Journal of Computational Physics}, 449:\penalty0 110768, 2022.

\bibitem[Lippe et~al.(2023)Lippe, Veeling, Perdikaris, Turner, and Brandstetter]{lippe2023pde}
Phillip Lippe, Bas Veeling, Paris Perdikaris, Richard Turner, and Johannes Brandstetter.
\newblock Pde-refiner: Achieving accurate long rollouts with neural pde solvers.
\newblock \emph{Advances in Neural Information Processing Systems}, 36:\penalty0 67398--67433, 2023.

\bibitem[Huang and Perdikaris(2025)]{huang2025physicscorrect}
Xinquan Huang and Paris Perdikaris.
\newblock Physicscorrect: A training-free approach for stable neural pde simulations.
\newblock \emph{arXiv preprint arXiv:2507.02227}, 2025.

\end{thebibliography}
\end{document}